\documentclass[lettersize,journal,twoside]{IEEEtran}
\usepackage{amsmath,amssymb,amsfonts}
\usepackage[ruled, vlined, linesnumbered, nofillcomment]{algorithm2e}
\SetKwFor{For}{for (}{) $\lbrace$}{$\rbrace$}

\SetCommentSty{mycommfont}
\usepackage[para,online,flushleft]{threeparttable}
\usepackage{array}
\usepackage{multirow}
\usepackage{makecell}
\usepackage{booktabs}
\usepackage[caption=false,font=footnotesize]{subfig}
\usepackage{textcomp}
\usepackage{stfloats}
\usepackage{url}
\usepackage{verbatim}
\usepackage{graphicx}
\usepackage{cite}
\usepackage{scalerel}
\usepackage{adjustbox}
\usepackage{tikz}
\usepackage{bbm}
\usepackage{bm}
\usetikzlibrary{svg.path}
\definecolor{orcidlogocol}{HTML}{A6CE39}
\tikzset{
    orcidlogo/.pic={
        \fill[orcidlogocol] svg{M256,128c0,70.7-57.3,128-128,128C57.3,256,0,198.7,0,128C0,57.3,57.3,0,128,0C198.7,0,256,57.3,256,128z};
        \fill[white] svg{M86.3,186.2H70.9V79.1h15.4v48.4V186.2z}
        svg{M108.9,79.1h41.6c39.6,0,57,28.3,57,53.6c0,27.5-21.5,53.6-56.8,53.6h-41.8V79.1z M124.3,172.4h24.5c34.9,0,42.9-26.5,42.9-39.7c0-21.5-13.7-39.7-43.7-39.7h-23.7V172.4z}
        svg{M88.7,56.8c0,5.5-4.5,10.1-10.1,10.1c-5.6,0-10.1-4.6-10.1-10.1c0-5.6,4.5-10.1,10.1-10.1C84.2,46.7,88.7,51.3,88.7,56.8z};
    }
}

\DeclareFontFamily{OT1}{pzc}{}
\DeclareFontShape{OT1}{pzc}{m}{it}{<-> s * [1.05] pzcmi7t}{}
\DeclareMathAlphabet{\mathpzc}{OT1}{pzc}{m}{it}

\newcommand\orcidicon[1]{\href{https://orcid.org/#1}{\mbox{\scalerel*{
                \begin{tikzpicture}[yscale=-1,transform shape]
                \pic{orcidlogo};
                \end{tikzpicture}
            }{|}}}}
        
\newcommand{\plcomment}[1]{{\color{black}#1}}

\usepackage[bookmarks=true,hidelinks]{hyperref}

\usepackage[utf8]{inputenc}
\DeclareUnicodeCharacter{2217}{*}
\newcommand{\eqn}{Eqn.}
\newcommand{\fig}{Fig.}
\newcommand{\alg}{Alg.}
\newcommand{\algline}{Line}
\newcommand{\alglines}{Lines}
\newcommand{\tab}{Table}

\newcommand{\rv}{\bm}
\newcommand{\dist}{\mathcal}
\newcommand{\gau}{\mathpzc}

\usepackage{background}
\usepackage{stackengine}
\setstackEOL{\\}
\setstackgap{L}{\normalbaselineskip}

\SetBgScale{1}
\SetBgContents{\normalsize\stackunder[37cm]{\Longstack{%
This article has been accepted for publication in IEEE Transactions on Robotics. This is the author's version which has not been fully edited and\\content may change prior to final publication. Citation information: DOI 10.1109/TRO.2023.3348305}}{%
		\rotatebox{0}{\textcopyright \hspace{1ex}2024 IEEE. Personal use is permitted, but republication/redistribution requires IEEE permission.
			See \url{https://www.ieee.org/publications/rights/index.html} for more information.}}}
\SetBgColor{black}
\SetBgAngle{0}
\SetBgOpacity{0.7}
\SetBgScale{0.7}

\begin{document}

\title{GMMap: Memory-Efficient Continuous Occupancy Map Using Gaussian Mixture Model}

\author{Peter~Zhi~Xuan~Li$^{\textsuperscript{\orcidicon{0000-0002-5260-4995}}}$,~\IEEEmembership{Student Member,~IEEE},
Sertac~Karaman$^{\textsuperscript{\orcidicon{0000-0002-2225-7275}}}$,~\IEEEmembership{Member,~IEEE},

\vspace{-3pt}and Vivienne~Sze$^{\textsuperscript{\orcidicon{0000-0003-4841-3990}}}$,~\IEEEmembership{Senior Member,~IEEE}
\thanks{This work was supported
	in part by the NSF RTML under Grant 1937501 and in part by the NSF CPS
	under Grant 1837212. (\emph{Corresponding author: Peter Zhi Xuan Li.})
	
	Authors are with the Massachusetts Institute of Technology, Cambridge, MA  02139,  USA. (email: \href{mailto: peterli@mit.edu}{peterli@mit.edu}, \href{mailto: sertac@mit.edu}{sertac@mit.edu}, \href{mailto: sze@mit.edu}{sze@mit.edu}).
	
	Digital Object Identifier 10.1109/TRO.2023.3348305}
}



\maketitle
\begin{abstract}
Energy consumption of memory accesses dominates the compute energy in energy-constrained robots which require a compact 3D map of the environment to achieve autonomy.
Recent mapping frameworks only focused on reducing the map size while incurring significant memory usage during map construction due to the multi-pass processing of each depth image.
In this work, we present a memory-efficient continuous occupancy map, named GMMap, that accurately models the 3D environment using a Gaussian mixture model (GMM).
Memory-efficient GMMap construction is enabled by the single-pass compression of depth images into local GMMs which are directly fused together into a globally-consistent map.
By extending Gaussian Mixture Regression to model unexplored regions, occupancy probability is directly computed from Gaussians.
Using a low-power ARM Cortex A57 CPU, GMMap can be constructed in real-time at up to 60 images per second.
Compared with prior works, GMMap maintains high accuracy while reducing the map size by at least 56\%, memory overhead by at least 88\%, DRAM access by at least 78\%, and energy consumption by at least 69\%.
Thus, GMMap enables real-time 3D mapping on energy-constrained robots.
\end{abstract}

\section{Introduction}
\IEEEPARstart{E}{nergy-constrained} microrobots could enable a wide variety of applications, from autonomous navigation, search and rescue, and space exploration~\cite{valavanis2015handbook}.
Due to the limited battery capacity onboard these robots, the amount of energy available for actuation (\emph{i.e.}, mechanical systems) and computation (\emph{i.e.}, executing algorithms) is extremely limited.
\plcomment{For actuation, researchers of these robots showcased mechanical systems that consume very low power (\emph{i.e.}, under 100 mW)~\cite{gtmab, chukewad2021robofly, wood2013flight, suhr2005biologically, keennon2012development}. 
Thus, a key remaining factor for enabling autonomy is the lack of energy-efficient algorithms.
}

During the execution of algorithms, the energy consumption of memory operations (\emph{e.g.}, reading and writing data stored in cache and DRAM) could dominate the total compute energy.
For instance, the energy required for accessing on-chip memory (\emph{e.g.}, cache) is more than an order-of-magnitude higher than that when performing a 32-bit multiplication~\cite{markhoro}.
The energy consumption of memory access increases with the size and distance of the memory from the processor.
Within the same chip, accessing a higher-level L2 cache (a few MBs) requires up to an order-of-magnitude more energy than lower-level L0 and L1 caches (a few KBs).
However, accessing data stored in a larger, off-chip memory such as DRAM (GBs of storage) requires more than two orders-of-magnitude higher energy than smaller, on-chip (local) CPU caches~\cite{markhoro}.
The memory (capacity) usage of an algorithm not only consists of output variables but also input and temporary variables allocated during computation.
Thus, algorithms designed for many robotics applications, especially the ones involving energy-constrained robots, should be \emph{memory efficient} such that: \emph{i)} the number of \emph{memory accesses} do not dominate; \emph{ii)} amount of \emph{memory (capacity) overhead} for storing input and temporary variables is small enough to remain in lower-level caches. 

\begin{figure}[t]
    \centering
  \subfloat[Incremental construction of the GMMap from a depth image $Z_t$ and pose $T_t$ obtained at time $t$. Each depth image is compressed into a local GMMap $\dist{G}_t$ which is then fused with the global GMMap $\dist{M}_{t\text{--}1}$.\label{gmmap_intro_a}]{%
       \includegraphics[width=0.46\columnwidth]{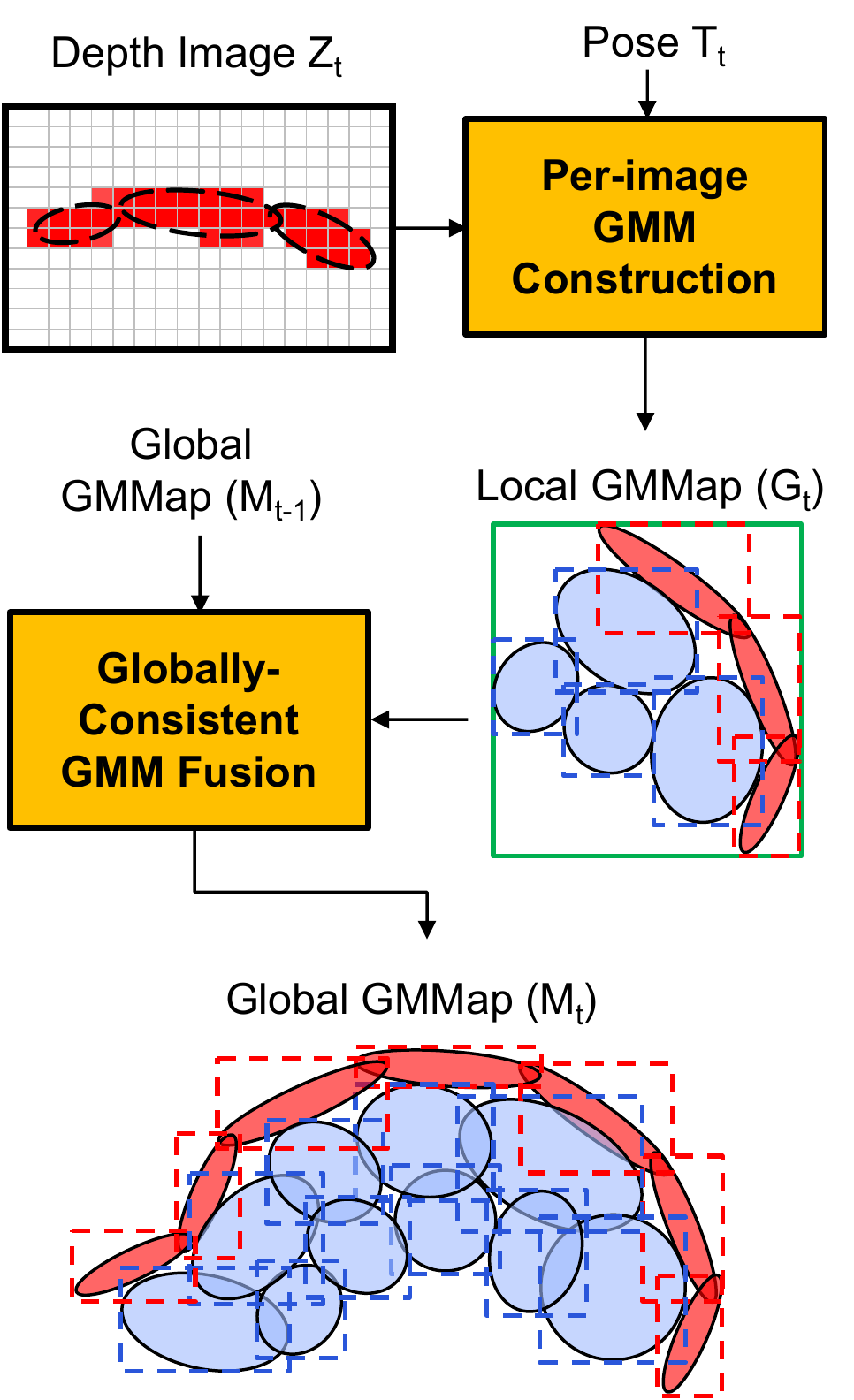}
       \label{fig:mapping_overview}
       }
  \hfill
  \subfloat[Visualization of the first floor of the MIT Stata Center and its GMMap representation consisting of GMMs representing occupied (red) and free (blue) regions. Each Gaussian is visualized as an ellipsoid in 3D.\label{gmmap_intro_b}]{%
        \includegraphics[width=0.48\columnwidth]{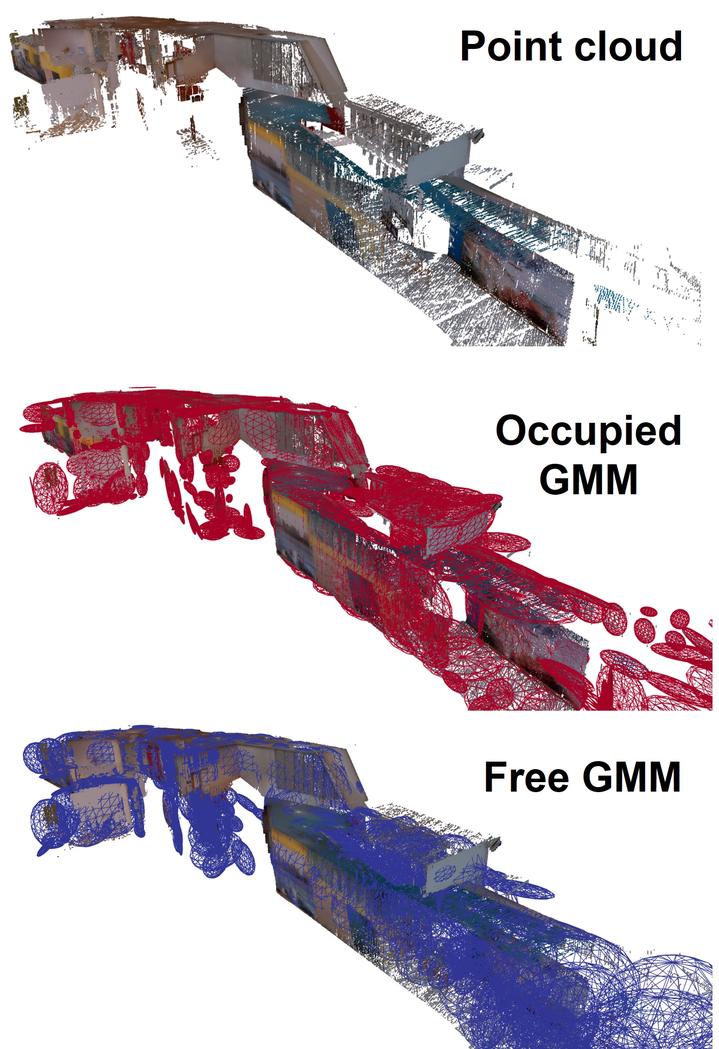}
        \label{fig:mapping_results}
        }
  \caption{Illustration of GMMap's \protect\subref{gmmap_intro_a} memory-efficient construction procedure and \protect\subref{gmmap_intro_b} representation for the MIT's Stata Center. 
  	\plcomment{Even though Gaussians are continuous and unbounded, GMMs representing occupied and free regions in the environment are represented by red and blue ellipsoids, respectively, created at a Mahalanobis distance of two for ease of visualization. The leaf nodes of the R-tree that store the GMMs are illustrated by dotted rectangles.}
  For the Stata Center, the GMMap models a continuous distribution of occupancy while requiring only 296KB to store.}
  \label{fig:gmmap_intro} 
\end{figure}


For mapping algorithms, both memory overhead and accesses could easily dominate.
During map construction, the multi-pass processing of sensor measurements requires them to be stored (\emph{i.e.}, as input and temporary variables) entirely in memory to support repeated accesses, which increases overhead and reduces the remaining memory for map storage.
Incrementally updating/reconstructing a previously-observed region in the map is typically performed by casting sensor measurement rays into the map.
Since these rays diverge away from the sensor origin, memory accesses along these rays often lack spatial and temporal locality required for effective cache usage, and thus require a significant number of memory accesses to DRAM.
Thus, achieving memory efficiency is both crucial and challenging for mapping algorithms.

In addition to achieving memory efficiency, the resulting map should satisfy the following requirements to enable memory-efficient, real-time processing of a variety of downstream applications that enable autonomy.
\begin{enumerate}
	 \item \textbf{Compactness}: A compact map can represent a larger portion of the environment in both on-chip (cache) and off-chip memory (DRAM). When accessing a region of the environment that does not reside in the cache, a compact map also reduces the number of energy-intensive DRAM accesses required to update the cache.
	
    \item \textbf{Modeling unexplored regions}: In autonomous exploration, the robot seeks to minimize the number of unexplored regions while traversing in obstacle-free regions. Thus, the ability to model unexplored regions enables state-of-the-art autonomous exploration algorithms based on frontier \cite{yamauchi1997frontier} or mutual information \cite{zhang2020fsmi, henderson2020efficient}.

    \item \textbf{Query compute efficiency}: During path planning and autonomous exploration, the robot needs to query multiple locations in the map to determine the current state of the environment~\cite{karaman2011sampling}. The results of these queries are often used to make decisions, such as the next location to travel, in real time. Thus, the state of the environment should be efficiently computed from the map.
\end{enumerate}

Current state-of-the-art mapping frameworks require the probabilistic modeling of \emph{occupancy} (\emph{i.e.}, whether or not an obstacle exists) at every location in the 3D environment. 
These frameworks can be classified based on their underlying probabilistic models used to infer occupancy.
For instance, the well-known framework, OctoMap~\cite{hornung2013octomap}, contains a set of Bernoulli random variables for modeling the occupancy at a discrete set of homogeneous regions in the environment. Even though OctoMap could model unexplored regions and achieve query efficiency, OctoMap is not compact enough for storage on energy-constrained robots. 
By using more compact models (\emph{e.g.}, set of Gaussians or kernels), recent frameworks (\emph{e.g.}, NDT-OM~\cite{saarinen20133d}, Hilbert Map~\cite{guizilini2018towards}, HGMM~\cite{srivastava2018efficient}) \plcomment{mostly} focused on reducing the map size while incurring significant memory overhead and accesses for multi-pass processing of raw sensor measurements in nearly every stage of the mapping pipeline.
In addition, the resulting maps produced by these frameworks cannot satisfy all of the above-mentioned requirements for enabling efficient downstream applications.

In this paper, we propose a continuous occupancy map comprised of a compact Gaussian mixture model (GMM), named GMMap, that is efficiently and accurately constructed from a sequence of depth images and poses of a robot.
To achieve significantly higher memory efficiency than prior works, our GMMap accurately compresses each depth image into a compact GMM in a \emph{single pass}, and \emph{directly} operates on Gaussians in the GMM (\emph{i.e.}, without other intermediate representations) for all remaining mapping operations. Our contributions are summarized as follows:
\begin{enumerate}
	\item \textbf{Single-pass compression}: A \emph{single-pass} procedure that accurately compresses a depth image into a local GMM for both free and occupied regions. Prior works~\cite{guizilini2018towards, srivastava2018efficient, eckart2016accelerated, o2018variable, dhawale2020efficient, goel2023probabilistic} require significant overhead for storing the entire image in memory due to \emph{multi-pass} processing.
	\item \textbf{Gaussian-direct map construction}: A novel procedure that \emph{directly} fuses the local GMMs across multiple images into a globally-consistent GMM without casting sensor rays (\emph{i.e.}, one ray for each pixel in the depth image) into the map. 
    Prior works~\cite{hornung2013octomap, Doherty2019, saarinen20133d, srivastava2018efficient} require a significant number of memory accesses during ray casting in order to update the previously-observed region that intersects with all sensor rays.
    
	\item \textbf{Gaussian-direct occupancy query}: An extension of Gaussian Mixture Regression to \emph{directly} compute occupancy from GMM while accounting for unexplored regions. Prior works require constructing and storing intermediate representations for modeling unexplored regions~\cite{guizilini2018towards, o2018variable} or do not model them at all~\cite{srivastava2018efficient}.
\end{enumerate}
%
%

In our previous work~\cite{spgf}, we proposed the Single-Pass Gaussian Fitting (SPGF) algorithm that enables single-pass compression of depth image into a GMM representing only the occupied region (\emph{i.e.}, a part of the first contribution) but not the obstacle-free region.
In this work, we not only extend our previous work to also construct a GMM representing the free region (\emph{i.e.}, the first contribution) but also illustrate how to directly operate on Gaussians during map construction and occupancy query (\emph{i.e.}, the second and third contribution).
An overview of the GMMap and its representation for the first floor of MIT's Stata Center is illustrated in \fig~\ref{fig:gmmap_intro}.

This paper is organized as follows. 
After analyzing existing works in Section~\ref{sec:related_work}, we describe how the occupancy is compactly represented and efficiently estimated from our GMMap in Section~\ref{sec:occupancy}. Memory-efficient algorithms that incrementally and accurately construct the GMMap given a sequence of depth images are presented in  Section~\ref{sec:construction}. Finally, we validate GMMap against existing works in terms of mapping accuracy, memory footprint, throughput, and energy consumption across multiple environments in Section~\ref{sec:results}.
\section{Related Work}~\label{sec:related_work}
Constructing an accurate and compact representation of the 3D environment is crucial for enabling many downstream robotics applications such as path planning and autonomous exploration. During the past few decades, many frameworks proposed different models to represent the distribution of the occupancy probability (\emph{i.e.}, the likelihood that a region contains an obstacle) across the 3D environment. These models exhibit different trade-offs in memory and computational efficiency during the construction and querying of the map.

\textbf{Discrete representations}: Some of the most popular mapping frameworks discretize the environment into cubic regions (\emph{i.e.}, grids in 2D and voxels in 3D) such that each region contains a Bernoulli random variable representing the occupancy probability and is assumed to be spatially independent of each other.
One of the earliest 2D mapping frameworks, the occupancy grid map~\cite{elfes1987sonar}, discretizes the environments into equally-sized grids. 
However, the map size is prohibitively large in 3D because the size scales cubically with the dimensions of the voxels and the environment.
To reduce map size in 3D, OctoMap~\cite{hornung2013octomap} and \plcomment{other voxel-based methods (\emph{e.g.}, \cite{funk2021multi, duberg2020ufomap})} store the occupancy probabilities in voxels whose sizes can adapt to homogeneous regions in the environment.
However, OctoMap and other voxel-based methods suffer from artifacts associated with voxelization and require a large amount of memory accesses during construction.
To incrementally construct the map given a set of sensor rays (more than 300,000 in each 640$\times$480 depth image), each ray is cast into the map to update the subset of voxels such rays intersect.
Since these rays diverge away from the sensor origin, memory accesses along these rays often lack spatial and temporal locality for effective cache usage (especially if the map is too large to fit in caches).
Since voxel-based methods are often not compact, updating the map requires a significant amount of memory accesses (more than 300,000 per image) to off-chip DRAM.
%

\textbf{Non-parametric representations}: To relax the spatial independence assumption in discrete map representations, Gaussian Process (GP) was proposed to estimate a continuous distribution of occupancy~\cite{o2012gaussian} using a covariance function that captures the spatial correlation among all sensor measurements. 
Since GP requires the storage of \emph{all} sensor measurements (since the beginning of the robotics experiment) to update the covariance function, the memory overhead scales with the total number of measurements $N$. During a map query, the covariance function generates a large matrix that requires $O(N^3)$ to invert, which greatly reduces the query efficiency. 
To enable faster map construction and query, recent non-parametric methods such as GPOctoMap~\cite{JWang-ICRA-16} and BGKOctoMap-L~\cite{Doherty2019} discretize the environment into blocks of octrees (\emph{i.e.}, a test-data octree).
For subsets of measurements (\emph{i.e.}, training data) that lie within each block, GPOctoMap and BGKOctoMap-L update the octrees in each block and its neighbors (\emph{i.e.}, extended blocks) using GP and Bayesian Generalized Kernel (BGK) inference, respectively.
Similar to OctoMap, both GPOctoMap and BGKOctoMap-L directly operate on sensor rays that are cast into the map during incremental construction and require significant memory accesses to DRAM.
In addition, both frameworks require the storage of training data for each block during map construction which incurs significantly larger memory overhead than OctoMap.

\textbf{Semi-parametric representations}: To create an extremely compact representation of the environment, several frameworks compress the sensor measurements using a set of parametric functions (\emph{e.g.}, Gaussians or other kernels) which are then used to infer occupancy.
One of the well-known semi-parametric representations is the Normal Distribution Transform Occupancy Map (NDT-OM)~\cite{saarinen20133d} that partitions the environment into large voxels such that measurements within each voxel are represented by a Gaussian.
Since measurements within a voxel could belong to multiple objects, representing them with a single Gaussian often leads to a loss of accuracy in the resulting map.
\plcomment{Due to the ineffective cache usage during the casting all sensor rays (\emph{i.e.}, more than 300,000) from each depth image into the map}, NDT-OM also requires significant memory access to DRAM during map construction.

To further reduce map size, recent frameworks, such as Hilbert Map (HM)~\cite{guizilini2018towards}, \plcomment{Fast Bayesian Hilbert Map (Fast-BHM)~\cite{zhi2019continuous}}, Variable Resolution GMM (VRGMM) map~\cite{o2018variable}, Hierarchical GMM (HGMM) map~\cite{srivastava2018efficient}, compress sensor rays into special kernels (in HM) or Gaussians (in VRGMM and HGMM).
Such compression is performed using techniques such as Quick-Means (QM)~\cite{guizilini2018towards}, Hierarchical Expectation-Maximization (H-EM)~\cite{eckart2016accelerated}, Region Growing (RG)~\cite{dhawale2020efficient}, Self-Organizing GMMs (SOGMM)~\cite{goel2023probabilistic}, and \plcomment{Integrated Hierarchical GMMs (IH-GMM)~\cite{gao2023integrated}}.
However, these techniques require significant memory overhead to store all sensor measurements (more than 300,000 pixels in a 640$\times$480 depth image) due to their \emph{multi-pass} processing. 
Even though the resulting maps are compact after compression, they either could not model unexplored regions (in HGMM), or require online training (for a logistic regression classifier in HM) and intermediate representations to model these regions (using Monte Carlo sampling to create an intermediate grid map in VRGMM).
Even though our GMMap is also classified as a semi-parametric representation, we can accurately construct and query the map directly using Gaussians (while preserving unexplored regions) to reduce memory overhead and accesses.

\section{Occupancy Representation \& Estimation}~\label{sec:occupancy}
In this section, we describe how to compactly model a continuous distribution of occupancy using a Gaussian mixture model (GMM) in the proposed GMMap.
In addition, we illustrate how to directly estimate the occupancy probability from Gaussians using Gaussian Mixture Regression (GMR) while accounting for the initial unknown state of the environment so that the unexplored regions are preserved.
\plcomment{Unless stated otherwise, matrices are denoted using regular uppercase letters, random variables are denoted using bold uppercase letters, and vectors/scalars are denoted using regular lowercase letters. 
In addition, a lower or upper calligraphic letter represents an individual Gaussian (\emph{i.e.}, parameterized by mean and covariance) or a set of Gaussians in a GMM (\emph{i.e.}, parameterized by a set of mean, covariance, and weights), respectively.
Depending on the context, we use these calligraphic letters to interchangeably represent either general Gaussian (or GMM) distributions or their instantiations with specific parameters.
}

Let $\rv{X} \in \mathbb{R}^3$ denote the 3D coordinate in the world frame. Let $\rv{O} \in \mathbb{R}$ denote the occupancy value such that regions with values greater than one are occupied with obstacles, and regions with values less than zero are obstacle free. 
In addition, unexplored regions have an occupancy value near 0.5.
Let $\rv{P}$ denote the joint random variable such that
\begin{equation} \label{eqn:joint_var}
	\rv{P} = \begin{bmatrix} 
		\rv{X} \\ \rv{O}
	\end{bmatrix}.
\end{equation}

The map $\dist{M}$ of the 3D environment is represented by the following GMM which is an \emph{unnormalized} distribution for the joint variable $\rv{P}$, \emph{i.e.}, 
\begin{equation}\label{eqn:joint_dist}
	\dist{M}_{\rv{P}}(p) \sim \sum_{i = 1}^{K}\pi_i\dist{N}\left(p \mid \mu_i, \Sigma_i\right),
\end{equation}
where $K$ is the number of Gaussians, and $\dist{N}(\cdot)$ is a Gaussian distribution.
The weight $\pi_i$, mean $\mu_i$, and covariance $\Sigma_i$ are the parameters of the $i$th Gaussian such that
\begin{equation} \label{eqn:gau_params}
	\mu_i = \begin{bmatrix} 
		\mu_{i\rv{X}} \\ \mu_{i\rv{O}} 
	\end{bmatrix}, \hspace{1ex}
	\Sigma_i = \begin{bmatrix} 
	\Sigma_{i\rv{X}} & \Sigma_{i\rv{XO}} \\ \Sigma_{i\rv{OX}} & \Sigma_{i\rv{O}}
	\end{bmatrix}.
\end{equation}
Note that the GMM in \eqn~\eqref{eqn:joint_dist} can be compactly stored because each Gaussian is parameterized by $\mu_{i}$, $\Sigma_i$, and $\pi_i$. For the rest of the paper, we drop the index $i$ for all variables when we refer to any Gaussian in the GMM.

During the experiment, the robot makes a sequence of range measurements. Each range measurement consists of a ray that originates from the robot, passes through a free region, and ends at the surface of an obstacle (occupied region).
Regions that are traversed by all \plcomment{measurement} rays are \emph{observed} by the robot.
We determine the parameters of the GMM in \eqn~\eqref{eqn:joint_dist} using range measurements such that it compactly models all observed regions. Thus, regions that have not been observed (\emph{i.e.}, unexplored) cannot be modeled by the GMM alone.

To compactly model the unexplored region, we use the \emph{unexplored prior} $\dist{Q}_{\rv{O} \mid \rv{X}}$ with its weight $\pi_0$ to represent the initial unknown state of the entire environment, \emph{i.e.},
\begin{equation}\label{eqn:prior_dist}
	\dist{Q}_{\rv{O} \mid \rv{X}}(o \mid x) = \dist{N}(o \mid \mu_0, \sigma_0^2),
\end{equation}
where 
\begin{equation}
	\mu_0 = 0.5, \hspace{1ex} \sigma_0^2 = 0.25.
\end{equation}
The weight $\pi_0$ should be set to a large value such that measurements from multiple timesteps are required to shift the occupancy value of an unexplored region (\emph{i.e.}, 0.5) towards zero (free region) or one (occupied region) during GMR.

Unlike prior semi-parametric representations that estimate occupancy probability using either a classifier that requires additional online training~\cite{ramos2016hilbert,guizilini2018towards} or intermediate representations that require additional memory overhead~\cite{o2018variable}, we efficiently preserve these regions by incorporating the unexplored prior into the Gaussian Mixture Regression (GMR)~\cite{sung2004gaussian}.
We describe the GMR procedure used to estimate occupancy directly from Gaussians as follows.

Using the GMM in \eqn~\eqref{eqn:joint_dist} and the unexplored prior in \eqn~\eqref{eqn:prior_dist}, the occupancy $\rv{O}$ conditioned on the query location $\rv{X} = x$ is computed as
\begin{equation}\label{eqn:cond_dist}
	\mathbb{P}_{\rv{O} \mid \rv{X}}(o \mid x) = \sum_{i = 0}^{K}\omega_i(x)\dist{N}\left(o \mid m_i(x), \sigma^2_i(x)\right),
\end{equation}
where
\begin{align}
	\omega_i(x) &= 
	\begin{cases}
		\frac{\pi_0}{\sum_{j = 1}^{K}\pi_j \dist{N}(x \mid \mu_{j\rv{X}}, \Sigma_{j\rv{X}}) + \pi_0},  &\text{if }i = 0,\vspace{1ex}\\
		\frac{\pi_i \dist{N}(x \mid \mu_{i\rv{X}}, \Sigma_{i\rv{X}})}{\sum_{j = 1}^{K}\pi_j \dist{N}(x \mid \mu_{j\rv{X}}, \Sigma_{j\rv{X}}) + \pi_0},  &\text{otherwise},
	\end{cases} \label{eqn:mixing_weight}\vspace{1ex}\\
	m_i(x) &= 
	\begin{cases}
		\mu_0,  &\text{if }i = 0,\\
		\mu_{i\rv{O}} + \Sigma_{i\rv{OX}}\Sigma^{-1}_{i\rv{X}}(x - \mu_{i\rv{X}}),  &\text{otherwise},
	\end{cases} \label{eqn:m_i}\vspace{1ex}\\
	\sigma^2_i(x) &= 
	\begin{cases}
		\sigma_0^2,  &\text{if }i = 0,\\
		\Sigma_{i\rv{O}} - \Sigma_{i\rv{OX}}\Sigma_{i\rv{X}}^{-1}\Sigma_{i\rv{XO}},  &\text{otherwise}.
	\end{cases} \label{eqn:sigma_i}
\end{align}

The expected occupancy value and its variance at location $x$ is regressed using GMR as
\begin{align}
	m(x) &= \mathbb{E}[\rv{O}|\rv{X}=x] = \sum_{i = 0}^{K}\omega_i(x)m_i(x), \label{eqn:occ_value} \\
	v(x) &= \text{Var}[\rv{O}|\rv{X}=x] = \sum_{i = 0}^{K}\omega_i(x)\left(m_i(x)^2 + \sigma^2_i(x)\right) \notag \\
	& \hspace{20ex} - m(x)^2. \label{eqn:occ_variance}
\end{align}

The occupancy value can transition suddenly across the boundaries separating occupied and free regions (\emph{e.g.}, at surfaces of obstacles). To better capture such transitions, each Gaussian in the GMMap models either an occupied or free region, but not both. 
Thus, the set of Gaussians representing occupied regions is defined as \emph{occupied Gaussians} with an occupancy value of \emph{one} (\emph{i.e.}, $\mu_{\rv{O}} = 1$).
In addition, the set of Gaussians representing free regions is defined as \emph{free Gaussians} with an occupancy value of \emph{zero} (\emph{i.e.}, $\mu_{\rv{O}} = 0$).

Representing occupied and free regions separately also guarantees that the expectation $m(x)$ in \eqn~\eqref{eqn:occ_value} is bounded within $[0,1]$. Thus, the expectation $m(x)$ becomes the \emph{occupancy probability} of the environment at the query location $x$.
In addition, the covariance terms $\Sigma_{\rv{XO}}$, $\Sigma_{\rv{OX}}$ and $\Sigma_{\rv{O}}$ in \eqn~\eqref{eqn:gau_params} become zero for all Gaussians, which significantly simplifies the entire GMR procedure and reduces the memory required to store the Gaussians in our map.

\plcomment{Even though occupied and free regions are modeled separately, the occupancy probability predicted by GMR in \eqn~\eqref{eqn:occ_value} is dependent on the relative contribution of all Gaussians because \emph{each Gaussian has no bound}. In particular, the relative contributions of occupied and free Gaussians in GMR provide a probabilistic way to resolve inconsistencies/noise of the sensor measurements, especially when obstacle and free Gaussians overlap near the obstacle surfaces.}
Although each Gaussian has no bound, its distribution tapers off from its mean at an exponential rate. Thus, the entire GMR procedure from \eqn~\eqref{eqn:cond_dist} to \eqref{eqn:occ_variance} can be accurately and efficiently approximated using a small subset of Gaussians whose Mahalanobis distances between their means to the query location $x$ are less than a threshold $\alpha_M$, \emph{i.e.}, 
\begin{equation}\label{eqn:maha_dist}
	\sqrt{(x-\mu_{\rv{X}})^\top\Sigma^{-1}_{\rv{X}}(x-\mu_{\rv{X}})} \leq \alpha_M,
\end{equation}
where $\mu_{X}$ and $\Sigma_{X}$ are parameters of a Gaussian defined in \eqn~\eqref{eqn:gau_params}. In our experiments, we chose $\alpha_M = 2$ to ensure that more than 95\% of the Gaussian distribution is considered.

To efficiently obtain the subset of Gaussians that satisfies \eqn~\eqref{eqn:maha_dist} in $O(\log(K))$ time (where $K$ is the total number of Gaussians), we store the GMMap using an R-tree~\cite{guttman1984r} constructed with bounding boxes that are axis-aligned with the world frame.
Since the surface that satisfies the equality in \eqn~\eqref{eqn:maha_dist} for each Gaussian can be visualized as an ellipsoid in 3D, the bounding box at the leaf node of the R-tree for each Gaussian is sized to enclose such ellipsoid.
In most figures, occupied and free Gaussians are represented by red and blue ellipsoids, respectively.
The occupied and free Gaussians in GMMap with their corresponding bounding boxes (dotted rectangles) are illustrated at the bottom of \fig~\ref{fig:mapping_overview}.
\section{Memory-Efficient Map Construction}~\label{sec:construction}
In this section, we present a memory-efficient framework to construct the GMMap $\dist{M}$ (\emph{i.e.}, \eqn~\eqref{eqn:joint_dist}). 
At each timestep $t$, we \emph{incrementally} constructs the GMMap $\dist{M}$ by updating the previous GMMap $\dist{M}_{t-1}$ with current measurements from the depth image $Z_t \in \mathbb{R}^{U \times V}$ obtained at pose $T_t \in \mathbb{SE}(3)$. As illustrated in \fig~\ref{fig:mapping_overview}, our framework consists of the following two procedures executed sequentially for each depth image:
\begin{enumerate}
	\item \textbf{Per-image GMM construction:} The depth image $Z_t$ with width $U$ and height $V$ obtained at pose $T_t$ is compressed into a compact local GMMap $\dist{G}_t$. 
	A memory-efficient algorithm is proposed in Section~\ref{subsec:per_img_gmm} to perform such compression one pixel at a time in a \emph{single pass} through the depth image.
	Memory overhead is greatly reduced by avoiding the storage of the entire depth image in memory which is required for prior multi-pass approaches~\cite{guizilini2018towards, srivastava2018efficient, eckart2016accelerated, o2018variable, dhawale2020efficient, goel2023probabilistic}.
	
	\item \textbf{Globally-consistent GMM fusion:} The local GMMap $\dist{G}_t$ is fused into the previous global GMMap $\dist{M}_{t-1}$ to obtain the updated GMMap $\dist{M}_t$.
	A memory-efficient algorithm is proposed in Section~\ref{subsec:global_gmm_fusion} to perform such fusion \emph{directly} using Gaussians.
	The amount of memory accesses is greatly reduced by avoiding casting rays (more than 300,000) from each 640$\times$480 image into the map as seen in prior works~\cite{hornung2013octomap, Doherty2019, saarinen20133d, srivastava2018efficient}.
	
\end{enumerate}

To efficiently update Gaussian parameters during the above-mentioned procedures with little memory overhead, we present preliminaries that illustrate \emph{in-place} construction of Gaussians using the method of moments (MoM)~\cite{scott2001kernels} in Section~\ref{subsec:gau_update_prelim}.

\subsection{Efficiently Updating Gaussian Parameters}~\label{subsec:gau_update_prelim}
We illustrate how to efficiently update the parameters of Gaussians in-place \plcomment{\emph{given} the correspondences between measurements and Gaussians.}
In the method of moments (MoM)~\cite{scott2001kernels}, the first and second moments of a Gaussian are intermediate representations of its mean and covariance, respectively.
Let $\rv{P}=[\rv{X}, \rv{O}]^\top$ denote the joint variable for the 3D coordinate $\rv{X}$ (with respect to the current sensor origin) and its occupancy $\rv{O}$.
The unnormalized first $m^{(1)} \in \mathbb{R}^4$ and second $M^{(2)} \in \mathrm{S}^{4}_{+}$ moments of each Gaussian are defined as
\begin{equation}\label{eqn:moments}
	m^{(1)} = \xi \mathbb{E}[\rv{P}], \hspace{1ex}M^{(2)} = \xi \mathbb{E}[\rv{P}^2],
\end{equation}
where $\xi$ is a normalization constant. Thus, the mean $\mu$ and covariance $\Sigma$ of each Gaussian defined in \eqn~\eqref{eqn:gau_params} can be recovered \emph{in-place} from the unnormalized moments as
\begin{equation}\label{eqn:mom2gau_param}
	\mu = \frac{1}{\xi}m^{(1)}, \hspace{1ex}\Sigma = \frac{1}{\xi}M^{(2)} - \mu\mu^\top.
\end{equation}

Unnormalized moments can be incrementally updated without relying on past measurements (which do not need to be stored in memory). 
Thus, during map construction, the moments for each Gaussian are stored instead of its mean and covariance.
Recall that from Section~\ref{sec:occupancy}, each measurement (ray) consists of a point (\emph{i.e.}, end of a ray representing the surface of an obstacle) and a line (\emph{i.e.}, from the sensor origin to the end of the ray representing free region). 
Fusing a \emph{point} $p = [x, 1]^\top \in \rv{P}$ into an occupied Gaussian is computed as
\begin{subequations}\label{eqn:fuse_pt_gau}
	\begin{align}
		m^{(1)} &\leftarrow  m^{(1)} + p, \\
		M^{(2)} &\leftarrow  M^{(2)} + pp^\top, \\
		\xi &\leftarrow \xi + 1.
	\end{align}
\end{subequations}

Fusing a \emph{line} from the sensor origin to the endpoint $p = [x, 0]^\top$ into a free Gaussian is computed as
\begin{subequations}\label{eqn:fuse_line_gau}
	\begin{align}
		m^{(1)} &\leftarrow  m^{(1)} + \frac{\|p\|}{2}p, \label{eqn:fuse_line_gau_m1}\\
		M^{(2)} &\leftarrow  M^{(2)} + \frac{\|p\|}{3}pp^\top, \label{eqn:fuse_line_gau_m2}\\
		\xi &\leftarrow \xi + \|p\|.
	\end{align}
\end{subequations}
Note that the second term on the right side of \eqn~\eqref{eqn:fuse_line_gau_m1} and \eqref{eqn:fuse_line_gau_m2} is the closed-from expression for the first and second moments of the line, respectively.
Our closed-form expression is accurate and more computationally efficient than prior works~\cite{srivastava2018efficient, guizilini2018towards} that approximate both moments by using points sampled at a fixed interval along the line.

When regressing occupancy using GMR, the unnormalized weight $\pi$ of each occupied or free Gaussian should represent the amount of occupied or free evidence in the region where such Gaussian resides.
For each free Gaussian, its weight $\pi$ equals to the total length of all line segments used during construction. When a line from sensor origin to $p=[x,0]^\top$ is fused into a free Gaussian, its weight $\pi$ is updated as
\begin{equation}~\label{eqn:weight_update}
	\pi \leftarrow \pi + \|p\|.
\end{equation}
To ensure that the occupancy regressed using GMR is meaningful, the weights for occupied Gaussians should have the same unit as those for the free Gaussians.
Thus, when a new endpoint $p$ is fused into an occupied Gaussian, its weight is also updated using \eqn~\eqref{eqn:weight_update}.

Lastly, the Gaussian containing the fusion of two occupied or free Gaussians indexed by $i$ and $j$ is computed as
\begin{subequations}\label{eqn:fuse_gau_gau}
	\begin{align}
		m^{(1)} &\leftarrow  m_i^{(1)} + m_j^{(1)}, \\
		M^{(2)} &\leftarrow  M_i^{(2)} + M_j^{(2)}, \\
		\xi &\leftarrow \xi_i + \xi_j, \\
		\pi &\leftarrow \pi_i + \pi_j.
	\end{align}
\end{subequations}

\subsection{Per-Image GMM Construction}~\label{subsec:per_img_gmm}
\begin{figure*}
	\centering
	\includegraphics[width=\linewidth]{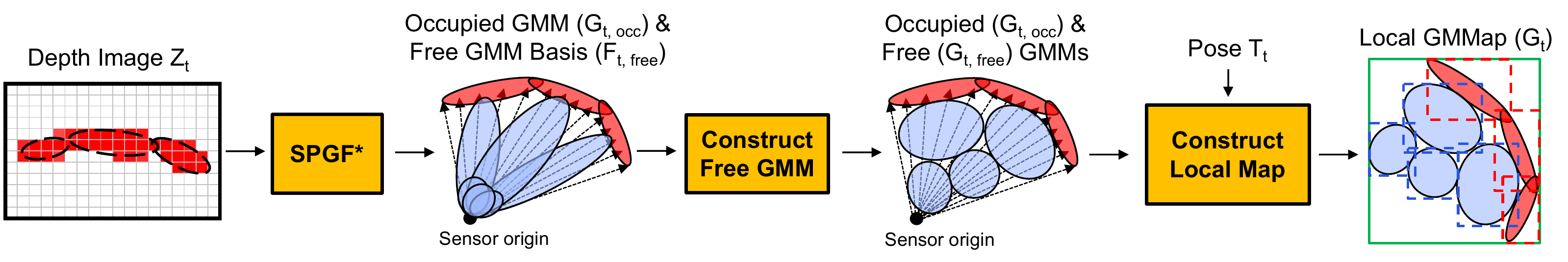}
	\caption{\textbf{Per-image GMM construction}: Constructing a local GMMap $\dist{G}_t$ that accurately represents both occupied and free regions from the current depth image $Z_t$ obtained at pose $T_t$. Rays associated with each pixel in the depth image are illustrated with dotted arrows. Occupied and free GMMs are illustrated with red and blue ellipsoids, respectively. Dotted rectangles in the map $\dist{G}_t$ represent the bounding boxes at the leaf nodes of the R-tree. The green rectangle represents the bounding box at the root node of the R-tree that encloses the entire map $\dist{G}_t$. }
	\label{fig:gmm_creation}
\end{figure*}

\begin{figure}[t!]
	\centering
	\includegraphics[width=0.6\columnwidth]{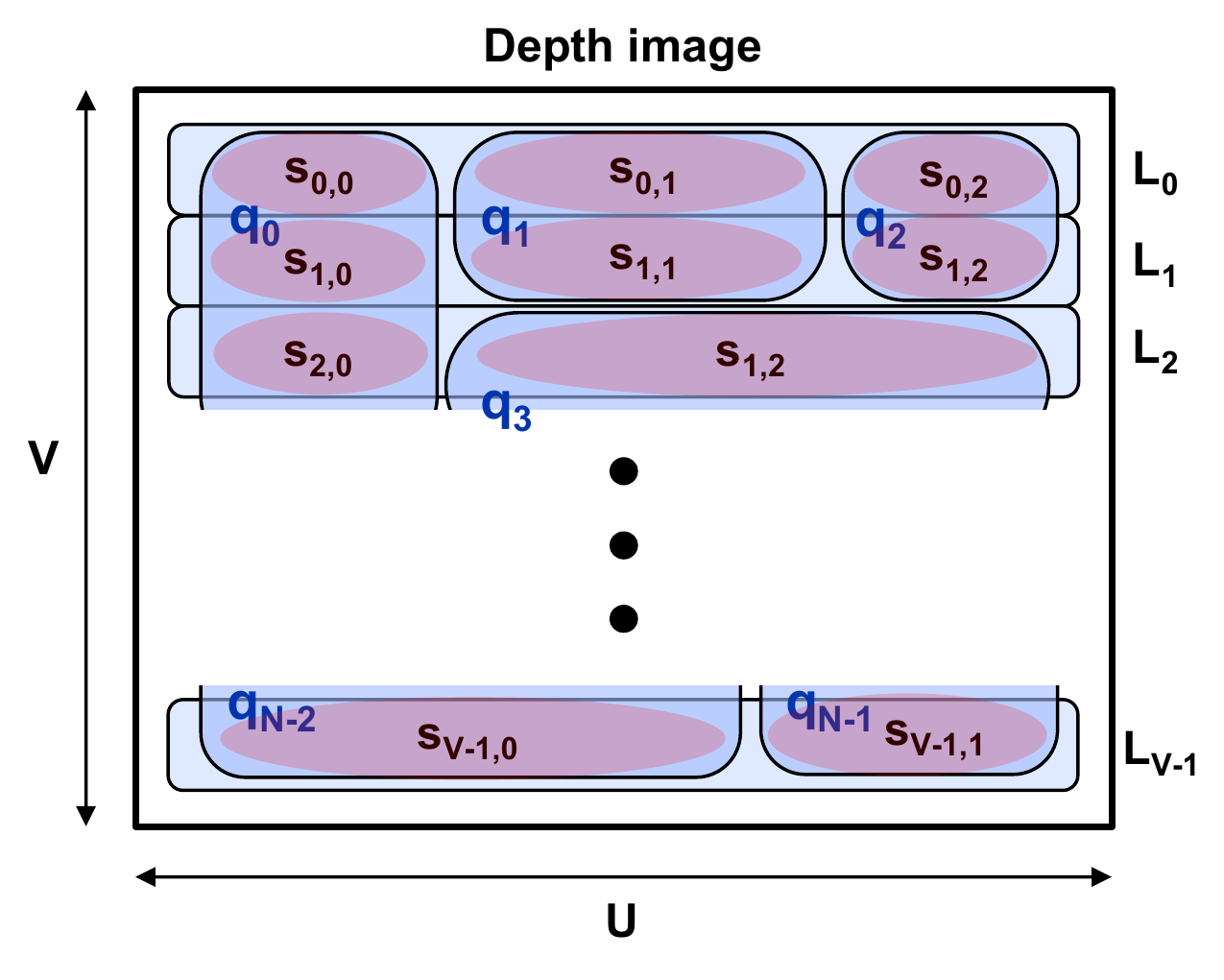}
	\caption{Single-pass processing of the depth image in SPGF* for constructing the set of Gaussians $\dist{Q}$ where each element $\gau{q}_j \in \dist{Q}$ contains one occupied Gaussian $\gau{a}_j$ and a free Gaussian basis $\gau{f}_j$. In Scanline Segmentation, each row (indexed by $v$) of the depth image is partitioned into a set of line segments $\dist{S} = \{\gau{s}_{v,i}\}$, which are fused across rows to form $\gau{q}_j \in \dist{Q}$ in Segment Fusion.}
	\label{fig:single_pass} 
\end{figure}

\begin{figure}
	\centering
	\subfloat[Visualization of the free Gaussian basis $\gau{f}_j = (\gau{\phi}_j, \gau{\beta}_j)$ associated with each occupied Gaussian $\gau{a}_j$. These bases cannot represent the free region faithfully (\emph{e.g.}, near the obstacles).\label{fig:free_space_a}]{%
		\includegraphics[width=0.47\columnwidth]{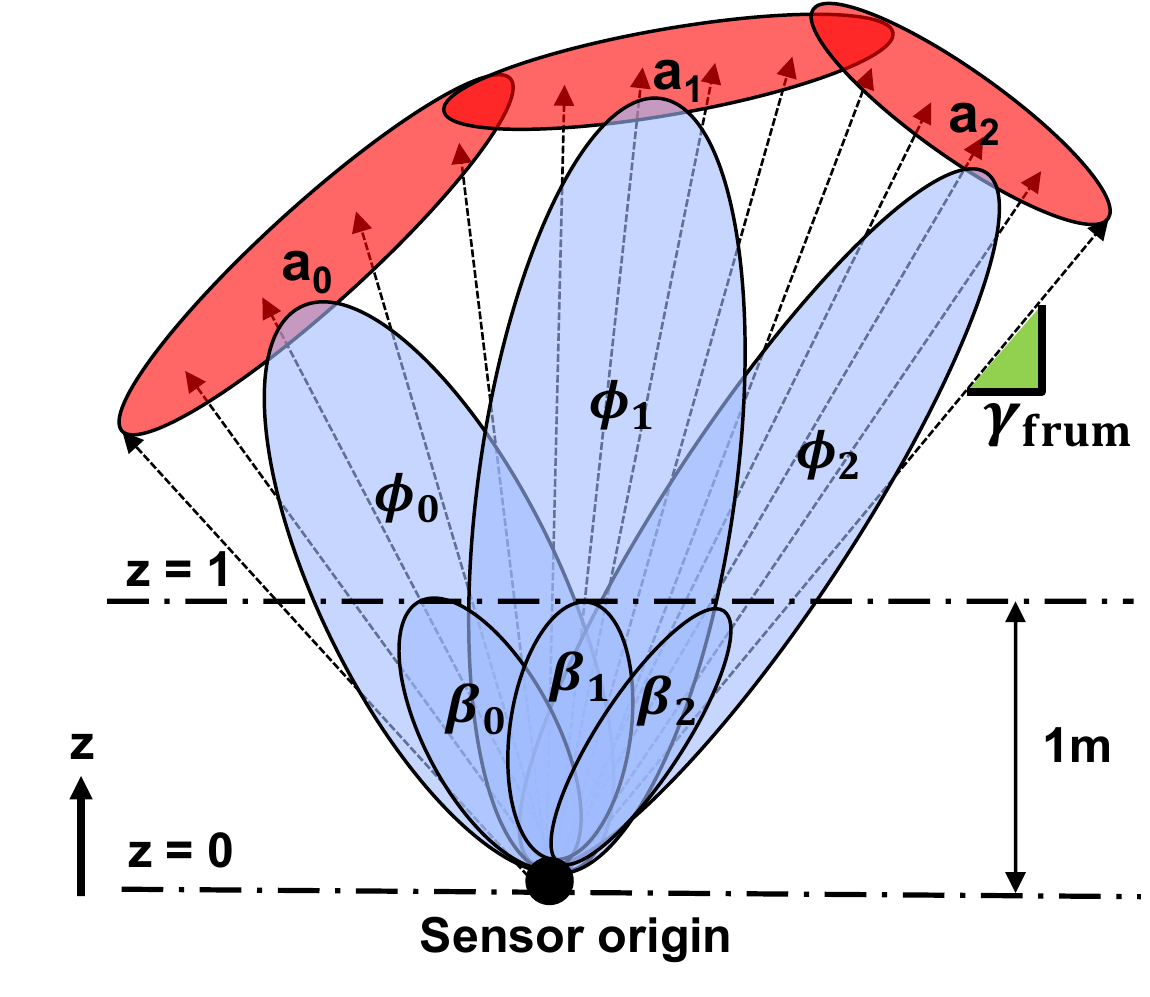}
		\label{fig:free_space_basis}
	}
	\hfill
	\subfloat[Visualization of the free Gaussian $\gau{b}_{i,j}$ that is recovered in subregion $\dist{B}_i$ from basis $\gau{f}_j = (\gau{\phi}_j, \gau{\beta}_j)$. The set of all free Gaussians accurately represents the free region. \label{fig:free_space_b}]{%
		\includegraphics[width=0.48\columnwidth]{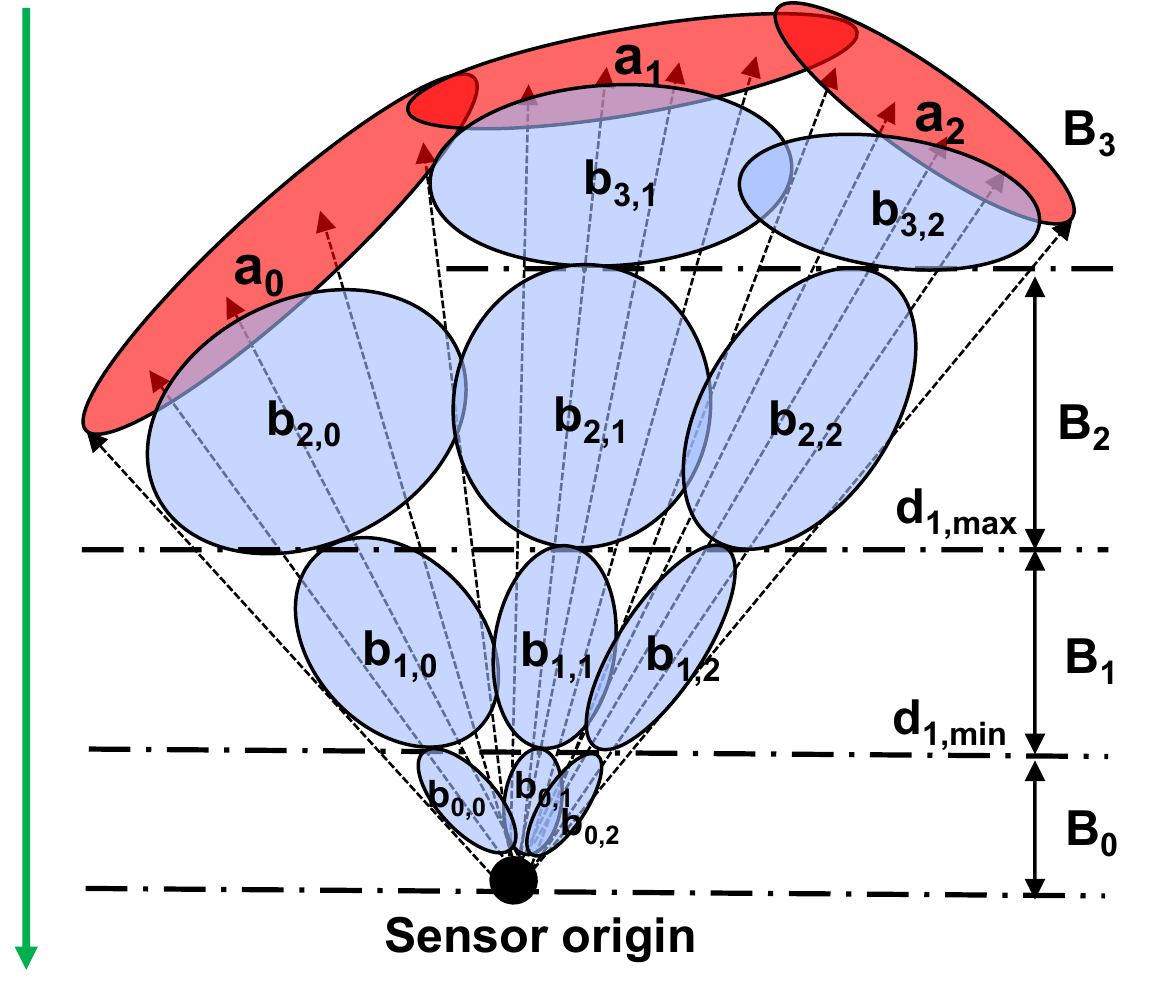}
		\label{fig:free_space_fusion}
	}
	\caption{Visualization of \protect\subref{fig:free_space_a} the free Gaussian basis which can be used to recover \protect\subref{fig:free_space_b} a corresponding set of free Gaussians in subregion $\dist{B}_i$ whose size increases with the distance from the sensor origin.}
	\label{fig:free_space} 
\end{figure}

As illustrated in \fig~\ref{fig:gmm_creation}, we present a single-pass algorithm that constructs a local GMMap $\dist{G}_t$ given the depth image $Z_t$ obtained at pose $T_t$. 
From Section~\ref{sec:occupancy}, occupied and free regions in the environment are separately modeled using occupied (visualized using red ellipsoids) and free (visualized using blue ellipsoids) Gaussians, respectively.
%
%
%
Thus, our algorithm, described in \alg~\ref{alg:gmm_creation}, creates both types of Gaussians in the local map $\dist{G}_t$ by executing the following procedures sequentially:

\begin{enumerate}
	\item \textbf{SPGF*} (\algline~\ref{per_img:spgf} in \alg~\ref{alg:gmm_creation}) : A memory-efficient algorithm that constructs the occupied GMM  $\dist{G}_{t, \text{occ}}$ and a compact free GMM basis $\dist{F}_{t, \text{free}}$ in a \emph{single pass} through the image $Z_t$ using only the endpoints of the sensor rays. 
	As illustrated in \fig~\ref{fig:single_pass}, operations within SPGF* are extended from our prior work SPGF~\cite{spgf}.

	\item \textbf{Construct Free GMM}  (\algline~\ref{per_img:free_gmm} in \alg~\ref{alg:gmm_creation}): By only considering the endpoints of the sensor rays in SPGF*, Gaussians represented by the free GMM basis $\dist{F}_{t, \text{free}}$ cannot represent free region encoded within the camera frustum very well (see \fig~\ref{fig:free_space_a}). Thus, the basis $\dist{F}_{t, \text{free}}$ is processed to construct the free GMM $\dist{G}_{t, \text{free}}$ that better represents the free region (see \fig~\ref{fig:free_space_b}).
	
	\item \textbf{Construct Local Map} (\alglines~\ref{per_img:gmm_concat} to \ref{per_img:build_rtree} in \alg~\ref{alg:gmm_creation}): Occupied $\dist{G}_{t, \text{occ}}$ and free $\dist{G}_{t, \text{free}}$ GMMs are transformed to the world frame using the pose $T_t$. Then, these GMMs are inserted into the R-tree to create the local map $\dist{G}_t$.
\end{enumerate}

\begin{algorithm}[t!]\small
	\caption{Per-Image GMM Construction}\label{alg:gmm_creation}
	\DontPrintSemicolon
	\KwIn{Depth image $Z_t$, pose $T_t$
	}
	\KwOut{Local GMMap $\dist{G}_t$}
	\SetKwBlock{Begin}{function}{end function}
	\Begin($\text{constructLocalGMM} {(} Z_t, T_t {)}$)
	{
		$\dist{G}_{t, \text{occ}}, \dist{F}_{t, \text{free}} \leftarrow \text{SPGF*}(Z_t)$ \label{per_img:spgf}\;
		$\dist{G}_{t, \text{free}} \leftarrow \text{constructFreeGMM} (\dist{F}_{t, \text{free}}) $ \label{per_img:free_gmm}\;
		$\dist{G}_t \leftarrow \dist{G}_{t, \text{free}} \cup \dist{G}_{t, \text{occ}}$ \label{per_img:gmm_concat}\;
		$\dist{G}_t \leftarrow \text{transform} (\dist{G}_t, T_t)$ \label{per_img:gmm_transform} \;
		$\dist{G}_t \leftarrow \text{constructRtree} (\dist{G}_t)$ \label{per_img:build_rtree}\;
		\Return{$\dist{G}_t$} 
	}
	
	\SetKwBlock{Begin}{subfunction}{end subfunction} 
	\Begin($\text{SPGF*} {(} Z_t {)}$ \label{per_img:spgf_start}) 
	{  
		$\dist{Q} \leftarrow \varnothing, \dist{Q}_{\text{prev}} \leftarrow \varnothing$  \;
		\For{$v = 0;\ v < V;\ v = v + 1$}{\label{spgf:parallel}
			$L_v \leftarrow \text{extractScanline}(Z_t, v)$ \label{spgf:scan_line} \;
			$\dist{S} \leftarrow \text{scanlineSegmentation}(L_v)$\; \label{spgf:scan_seg}
			\eIf{$v = 0$}{
				$\dist{Q}_{\text{prev}} \leftarrow \dist{S}$\;
			}{
				$\dist{Q}_{\text{prev}}$, $\dist{Q}_{\text{comp}} \leftarrow \text{segmentFusion}(\dist{Q}_{\text{prev}}, \dist{S})$\; \label{spgf:seg_fuse}
				$\dist{Q} \leftarrow \dist{Q} \cup \dist{Q}_{\text{comp}}$\; \label{spgf:g_append}
			}
		}
		$\dist{Q} \leftarrow \dist{Q} \cup \dist{Q}_{\text{prev}}$\;
		$\dist{G}_{t, \textup{occ}}, \dist{F}_{t, \textup{free}} \leftarrow \dist{Q}$\;
		\Return{$\dist{G}_{t, \textup{occ}}, \dist{F}_{t, \textup{free}}$}  \label{per_img:spgf_end}
	}
\end{algorithm}

\textbf{SPGF*} (\alglines~\ref{per_img:spgf_start} to \ref{per_img:spgf_end} in \alg~\ref{alg:gmm_creation}): 
The SPGF* algorithm constructs an occupied GMM  $\dist{G}_{t, \text{occ}}$ and a compact free GMM basis $\dist{F}_{t, \text{free}}$ by processing one scanline (\emph{i.e.}, a row of pixels) at a time in a \emph{single pass} through the entire depth image $Z_t$, which avoids the storage of the entire depth image in memory required for prior \emph{multi-pass} approaches\cite{guizilini2018towards, srivastava2018efficient, eckart2016accelerated, o2018variable, dhawale2020efficient, goel2023probabilistic}.
The SPGF* algorithm is a direct extension of our prior work SPGF~\cite{spgf} which exploits the intrinsic properties of the depth camera to accurately infer the correspondences between Gaussians and depth measurements.
We briefly summarize SPGF for creating occupied Gaussians as follows.

Recall that each depth image contains a set of \emph{organized} depth measurements such that measurements that are neighbors in the image are likely neighbors on the same planar obstacle surface.
\plcomment{In SPGF, we derive distance-based thresholds to accurately determine which subset of neighboring measurements belong to the same planar surface (and Gaussian) across scanlines (\emph{i.e.}, rows) of each image.}
As illustrated in \fig~\ref{fig:single_pass}, each scanline is denoted by $L_v$, where $v$ is the row index. In Scanline Segmentation (SS, \algline~\ref{spgf:scan_seg}), pixels from each scanline are partitioned into a set of line segments $\dist{S}$ such that each segment $\gau{s} \in\dist{S}$ represents a locally planar surface with distinct orientation.
\plcomment{Specifically, neighboring pixels belong to the same line segment if the distances among them fall below an adaptive threshold derived from the orientation of the surface and camera parameters.}
In Segment Fusion (SF, \algline~\ref{spgf:seg_fuse}), segments are fused across successive scanlines to form a set of completed Gaussians $\dist{Q}_{\text{comp}}$ (appended to output in \algline~\ref{spgf:g_append}) and incomplete Gaussians $\dist{Q}_{\text{prev}}$ (for fusion with the next scanline). 
\plcomment{Specifically, segments across the neighbor scanlines belong to the same surface (and Gaussian) if they are sufficiently close and parallel with each other.}


%
%

To create a free Gaussian basis for each occupied Gaussian, the implementation of SS and SF in SPGF* is almost identical to those in SPGF except for the following differences.
Recall that each pixel in the depth image is a sensor ray that originates from the robot.
In SPGF, only occupied GMM $\dist{G}_{t, \text{occ}}$ is constructed using endpoints of the sensor rays from all depth pixels in the image. In particular, \eqn~\eqref{eqn:fuse_pt_gau} and \eqref{eqn:fuse_gau_gau} are used to construct each occupied Gaussian (say $\gau{a}_j$) in SS and SF, respectively.
Since we would like to construct Gaussians associated with the free region as well, SPGF* constructs two free Gaussians $\gau{\phi}_j$ (using the entire sensor ray) and $\gau{\beta}_j$ (using normalized sensor ray with depth $z = 1$) concurrently with each occupied Gaussian $\gau{a}_j$.
Those free Gaussians are constructed using \eqn~\eqref{eqn:fuse_line_gau} in SS and \eqn~\eqref{eqn:fuse_gau_gau} in SF. Thus, for SPGF*, each element in the set $\dist{S}$, $\dist{Q}$, $\dist{Q}_{\text{prev}}$ and $\dist{Q}_{\text{comp}}$ of \alg~\ref{alg:gmm_creation} includes the occupied Gaussian $\gau{a}_j$ with its associated free Gaussians $\gau{\phi}_j$ and $\gau{\beta}_j$.

\fig~\ref{fig:free_space_a} illustrates the free Gaussians $\gau{\phi}_j$ and $\gau{\beta}_j$ associated with each occupied Gaussian $\gau{a}_j$. Note that the free Gaussians $\gau{\phi}_j$ and $\gau{\beta}_j$ do not represent the free space traversed by the sensor rays very well. However, these free Gaussians can be used to reconstruct a better representation of free space as illustrated in \fig~\ref{fig:free_space_b} during the subsequent procedure. Thus, we define each free Gaussian basis $\gau{f}_j$ such that $\gau{f}_j = (\gau{\phi}_j, \gau{\beta}_j)$. The set of all free Gaussian bases generated at the output of SPGF* forms the free GMM basis $\dist{F}_{t, \text{free}}$.

Since the criteria for constructing and updating the Gaussians in SPGF* is identical to SPGF, SPGF* inherits many desirable properties from SPGF.
Since SS in \algline~\ref{spgf:scan_seg} dominates SPGF* and can be executed independently for each scanline, SPGF* can be parallelized by concurrently executing SS for different scanlines across multiple CPU or GPU cores. 
Due to single-pass pixel-per-pixel processing in SS, only one pixel is stored in memory at any time. Thus, SPGF* is memory-efficient and avoids the storage of the entire depth image in memory as seen in most prior works.

\textbf{Construct Free GMM} (\alg~\ref{alg:free_gmm}): In this section, we present \alg~\ref{alg:free_gmm} that directly generates the set of free Gaussians $\dist{G}_{t, \text{free}}$ from their basis $\dist{F}_{t, \text{free}}$. These Gaussians should accurately and compactly model the free space traversed by the sensor rays (\emph{i.e.}, within the viewing frustum).
In prior works~\cite{guizilini2018towards, srivastava2018efficient}, the free Gaussians $\dist{G}_{t, \text{free}}$ are inefficiently constructed from a large number of free-space points sampled at a fixed interval along all sensor rays.
In contrast, the free Gaussians $\dist{G}_{t, \text{free}}$ in GMMap are directly constructed from their basis $\dist{F}_{t, \text{free}}$ with little computational and memory overhead.

The free space is contained within the viewing frustum which is a pyramidal region with significantly different symmetries than each elliptical equipotential surface of the Gaussian distribution.
Thus, as illustrated in \fig~\ref{fig:free_space_a}, free Gaussians (\emph{i.e.}, $\phi$ and $\beta$) from the basis $\dist{F}_{t, \text{free}}$ cannot faithfully represent the free region (especially near the obstacles). 
To achieve a more accurate representation, we partition the viewing frustum into subregions $\{\dist{B}_0, \dist{B}_1, \dots\}$ along the $z$-axis that is perpendicular to the image plane of the camera.
Each subregion $\dist{B}_i$ is enclosed between two partitioning planes $z = d_{i, \text{max}}$ and $z = d_{i, \text{min}}$.
As illustrated in \fig~\ref{fig:free_space_fusion}, free Gaussians are constructed to model each subregion separately.

The free Gaussians in each subregion (\fig~\ref{fig:free_space_b}) can be directly recovered from each basis $\gau{f} = (\phi, \beta)$.
Let the index $i_{\gau{f}}$ denote the \emph{minimum index} across all subregions containing the endpoints of rays used to construct Gaussian $\phi$. For instance, the index $i_{\gau{f}} = 2$ for basis $\gau{f}_0 = (\phi_0, \beta_0)$ in \fig~\ref{fig:free_space_fusion}. 
The subregion $\dist{B}_{i_{\gau{f}}}$ is the difference between the region from sensor origin to the obstacle (represented by Gaussian $\phi$) and the region from the sensor origin to the partitioning plane $d_{i_{\gau{f}}, \text{min}}$ (represented by Gaussian $\beta$ scaled to $d_{i_{\gau{f}}, \text{min}}$).
Each remaining subregion is the difference between regions from the sensor origin to two enclosing partitioning planes $d_{i, \text{min}}$ and $d_{i, \text{max}}$ (represented by Gaussian $\beta$ scaled to $d_{i, \text{min}}$ and $d_{i, \text{max}}$).
For each basis $\gau{f} = (\phi, \beta)$, the parameters of the free Gaussian $g$ (\emph{i.e.}, first moment $m^{(1)}_{\gau{g}}$, second moment $M^{(2)}_{\gau{g}}$, normalizing constant $\xi_{\gau{g}}$, and weight $\pi_{\gau{g}}$) in subregion $\dist{B}_i$ are directly recovered from the parameters of each basis as follows:
\begin{subequations}\label{eqn:recover_free_gau}
	\begin{align}
		m^{(1)}_{\gau{g}} &= 		
		\begin{cases}
			m^{(1)}_{\phi} - d^2_{i, \text{min}} m^{(1)}_{\beta}, &\quad \text{if } i = i_{\gau{f}}, \\[1ex]
			m^{(1)}_{\beta} \left( d^2_{i, \text{max}} - d^2_{i, \text{min}} \right), &\quad \text{if } 0 \leq i < i_{\gau{f}},\\
		\end{cases} \\
		M^{(2)}_{\gau{g}} &= 		
		\begin{cases}
			M^{(2)}_{\phi} - d^3_{i, \text{min}} M^{(2)}_{\beta}, &\quad \text{if } i = i_{\gau{f}}, \\[1ex]
			M^{(2)}_{\beta} \left( d^3_{i, \text{max}} - d^3_{i, \text{min}} \right), &\quad \text{if } 0 \leq i < i_{\gau{f}},\\
		\end{cases} \\
		\xi_{\gau{g}} =\hspace{2pt}& \pi_{\gau{g}} = 		
		\begin{cases}
			\xi_{\phi} - d_{i, \text{min}} \xi_{\beta}, &\quad \text{if } i = i_{\gau{f}}, \\[1ex]
			\xi_{\beta} \left( d_{i, \text{max}} - d_{i, \text{min}} \right), &\quad \text{if } 0 \leq i < i_{\gau{f}}.\\
		\end{cases} 
	\end{align}
\end{subequations}
See \fig~\ref{fig:free_space_fusion} for an illustration of the recovered free Gaussians $\dist{B}_{i,j}$ in subregion $\dist{B}_i$ generated from basis $\gau{f}_j$.

To retain high mapping fidelity, each subregion is sized according to its spatial resolution (\emph{i.e.}, the density of the sensor rays) such that regions with higher resolution are modeled by smaller Gaussians.
Since the sensor rays emanate outwards from the origin, the spatial resolution of each subregion $\dist{B}_i$ decreases as its index $i$ increases (see \fig~\ref{fig:free_space_fusion}).
%
%
To ensure that the maximum size of each Gaussian is inversely proportional to the spatial resolution, the distance between the partitioning planes that enclose each subregion $\dist{B}_i$ should increase with index $i$.
Thus, given the maximum slope of the frustum's boundary $\gamma_{\text{frum}}$ along the $z$ axis (see \fig~\ref{fig:free_space_basis}) and the initial distance $d_0$ between partitioning planes, the locations of these planes for each subregion $\dist{B}_i$ are computed as 
\begin{subequations}\label{eqn:partition_plane}
	\begin{align}
		d_{i, \text{max}} &= d_0\sum_{k = 0}^{k=i} (1 + \alpha_d \gamma_{\text{frum}})^{i} 
            = \frac{d_0\left((1 + \alpha_d \gamma_{\text{frum}})^{i+1} - 1\right)}{\alpha_d \gamma_{\text{frum}}},  \\
		d_{i, \text{min}} &= 
		\begin{cases}
			0, &\quad \text{if } i = 0, \\
			d_{i-1, \text{max}}, &\quad \text{otherwise},\\
		\end{cases}
	\end{align}
\end{subequations}
where $\alpha_d$ is a scaling parameter. We chose $\alpha_d = 0.5$ in all our experiments.

\begin{figure}
	\centering
	\subfloat[Fusion of two completely overlapping Gaussians (blue) into a single Gaussian (green).\label{fusion_scenario_a}]{%
		\makebox[0.45\columnwidth][c]{\includegraphics[width=0.35\columnwidth]
			{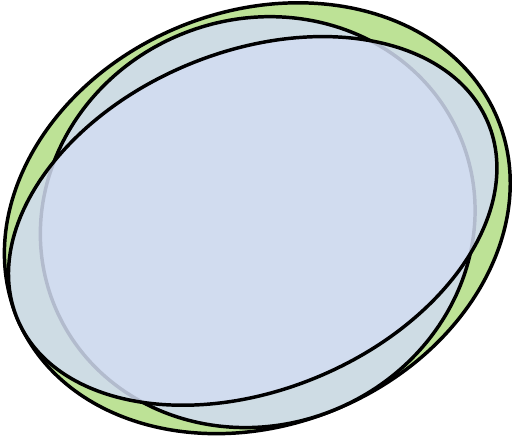}}  
		\label{fig:overlap_fusion}
	}
	\hfill
	\subfloat[Fusion of two partially overlapping Gaussians (blue) into a single Gaussian (green).\label{fusion_scenario_b}]{%
		\makebox[0.45\columnwidth][c]{\includegraphics[width=0.32\columnwidth]
			{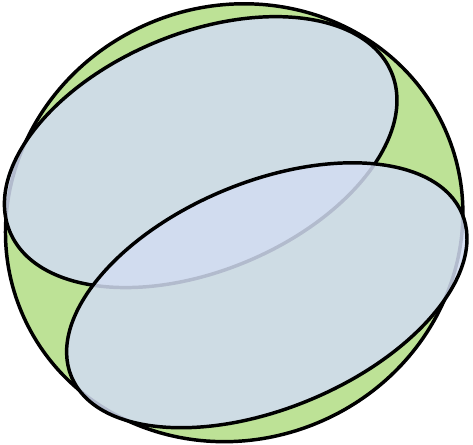}}  
		\label{fig:neighbor_fusion}
	}	
	\caption{Our fusion criteria using the Unscented Hellinger distance~\cite{kristan2010multivariate} allows for the creation of a single Gaussian (green) from two Gaussians (blue) when they \protect\subref{fusion_scenario_a} completely overlap to represent the same region or \protect\subref{fusion_scenario_b} partially overlap to represent neighboring parts of the same region in the environment.}
	\label{fig:fusion_scenario} 
\end{figure}

\begin{algorithm}[t!]\small
	\caption{Free GMM Construction From Basis} \label{alg:free_gmm}
	\DontPrintSemicolon
	\KwIn{Free GMM basis $\dist{F}_{t, \text{free}}$}
	\KwOut{Free GMM $\dist{G}_{t, \textup{free}}$}
	\SetKwBlock{Begin}{function}{end function} 
	\Begin($\text{constructFreeGMM} {(} \dist{F}_{t, \text{free}} {)}$)
	{
		$\dist{B} \leftarrow \varnothing, \dist{G}_{t, \textup{free}} \leftarrow \varnothing$ \;
		$i_{\text{max}} \leftarrow 0$ \;
		
		\tcp{Sort basis using its subregion index $i_{\gau{f}}$}
		\ForEach{$\gau{f} \in \dist{F}_{t, \textup{free}}$}{\label{free_gmm:sort_start}
			$i_{\gau{f}} \leftarrow \text{region}(\gau{f})$ \label{free_gmm:region_idx}\;
			$\dist{F}_{t, \text{free}} \leftarrow \dist{F}_{t, \text{free}} \setminus \gau{f}$ \;
			$\dist{B}_{i_{\gau{f}}} \leftarrow \dist{B}_{i_{\gau{f}}} \cup \gau{f}$ \label{free_gmm:sort_transfer}\;
			$i_{\text{max}} \leftarrow \max(i_{\text{max}}, i_{\gau{f}})$\; \label{free_gmm:sort_end}
		}
		
		\For{$i = i_{\text{max}};\ i \geq 0;\ i = i - 1$}{\label{free_gmm:creation_start}
			
			\tcp{See \eqn~\eqref{eqn:partition_plane}}
			$d_{i,\text{min}}, d_{i,\text{max}} \leftarrow \text{computePartitioningPlanes}(i)$ \label{free_gmm:planes} \;
			$\dist{Q} \leftarrow \varnothing$ \;
			
			\tcp{Recover free Gaussians from bases}
			\ForEach{$\gau{f} \in \dist{B}_i$}{\label{free_gmm:recover_start}
				\tcp{See \eqn~\eqref{eqn:recover_free_gau}} 
				$\gau{g} \leftarrow \text{recoverFreeGaussian}(\gau{f}, d_{i,\text{min}}, d_{i,\text{max}})$ \label{free_gmm:recover_free_gau}\;
				$\dist{B}_i \leftarrow \dist{B}_i \setminus \gau{f}$ \;
				\tcp{Each tuple $\gau{q}$ contains a Gaussian $\gau{q_{g}}$ generated from its basis $\gau{q_{f}}$}
				$\gau{q} \leftarrow (\gau{q_{g}} \leftarrow \gau{g}, \gau{q_{f} \leftarrow \gau{f}})$ \;
				$\dist{Q} \leftarrow \dist{Q} \cup \gau{q}$
			}\label{free_gmm:recover_end}
			
			\tcp{Fuse free Gaussians in subregion $\dist{B}_i$}
			\While{$\textup{notEmpty}(\dist{Q})$}{\label{free_gmm:fusion_start}
				$\gau{q} \leftarrow \text{front}(\dist{Q}) $ \label{free_gmm:queue_front}\;
				$\dist{C} \leftarrow \text{findNeighbors}(\dist{Q}, \gau{q})$ \label{free_gmm:neighbor} \;
				$\text{isFused} \leftarrow \text{false}$ \;
				
				\ForEach{$\gau{c} \in \dist{C}$}{
					\tcp{See \eqn~\eqref{eqn:fuse_gau_gau}}
					$\gau{r} \leftarrow \text{fuseGaussianAndBasis}(\gau{c}, \gau{q})$ \label{free_gmm:fuse_components}\;
					$d_h \leftarrow \text{unscentHellingerDistance}(\gau{r_g}, \gau{c_g}, \gau{q_{g}})$ \label{free_gmm:hell_dist}\;
					$s_r \leftarrow \text{geometricSimilarity}(\gau{c_g}, \gau{q_{g}})$ \label{free_gmm:geo_sim}\;
					\If{$d_h \leq s_r \cdot \alpha_{h, \text{free}}$}{ \label{free_gmm:sim_const}
						$\gau{q} \leftarrow \gau{r}$ \label{free_gmm:gau_fusion}\;
						$\dist{Q} \leftarrow \dist{Q} \setminus \gau{c}$ \label{free_gmm:neighbor_remove}\;
						$\text{isFused} \leftarrow \text{true}$ \;
					}
				}
				
				\If{$\textup{isFused} = \textup{false}$}{
					$\dist{Q} \leftarrow \dist{Q} \setminus \gau{q}$ \label{free_gmm:q_remove}\;
					$\dist{G}_{t, \textup{free}} \leftarrow \dist{G}_{t, \textup{free}} \cup \gau{q_{g}}$ \label{free_gmm:transfer_output}\;
					
					\tcp{Propagate fusion decision to subregion $\dist{B}_{i\text{-}1}$}
					\If{$i > 0$}{
						$\dist{B}_{i\text{--}1} \leftarrow \dist{B}_{i\text{--}1} \cup \gau{q_{f}}$ \label{free_gmm:transfer_subregion}\;
					}
				}
			}\label{free_gmm:fusion_end}
		}
		\Return{$\dist{G}_{t, \textup{free}}$}
	}
\end{algorithm}

Although free Gaussians recovered from the basis can accurately model the free region, they are not as compact as the occupied GMM $\dist{G}_{t,\text{occ}}$.
Thus, after recovery, free Gaussians are fused with each other in each subregion to further enhance the compactness of the map.
\alg~\ref{alg:free_gmm} efficiently performs Gaussian recovery and fusion.
After sorting each basis $\gau{f}$ into its associated subregion based on index $i_{\gau{f}}$ (\algline~\ref{free_gmm:sort_start} to \ref{free_gmm:sort_end}), free Gaussians are constructed within each subregion (from \algline~\ref{free_gmm:creation_start} onward) starting from the one that is furthermost away from the sensor origin (see the green arrow in \fig~\ref{fig:free_space_fusion}).
Using the bases, free Gaussians in each subregion $\dist{B}_i$ are initially recovered (\algline~\ref{free_gmm:recover_start} to \ref{free_gmm:recover_end}) and then fused with each other \plcomment{one pair at a time} in a region-growing approach (\algline~\ref{free_gmm:fusion_start} to \ref{free_gmm:fusion_end}).

During region growing, we need to ensure that the fused Gaussian can still accurately represent the free region within each subregion $\dist{B}_i$.
After fusing a free Gaussian $\gau{q_{g}}$ with its neighbor $\gau{c_g}$ (determined by whether their bounding boxes intersect in \algline~\ref{free_gmm:neighbor}), we accept the fused Gaussian $\gau{r_g}$ (in \algline~\ref{free_gmm:gau_fusion}) if it accurately represents its original components (\emph{i.e.}, $\gau{q_{g}}$ and $\gau{c_g}$).
In prior works~\cite{guizilini2018towards}, the fused Gaussian $\gau{r_g}$ is accepted if the probabilistic distance $d_h$ between two components $\gau{q_{g}}$ and $\gau{c_g}$ are below a pre-defined low threshold $\alpha_{h, \text{free}}$.
Thus, only Gaussians that completely overlap the same region can be fused (see \fig~\ref{fig:overlap_fusion}).
However, there exist many opportunities to fuse Gaussians that only partially overlap but accurately represent neighboring parts of the same region (see \fig~\ref{fig:neighbor_fusion}).
To also exploit these opportunities, our distance measure $d_h$ is computed between the fused Gaussians $\gau{r_g}$ and its components $\{\gau{q_{g}}, c_g\}$ using the Unscented Hellinger distance~\cite{kristan2010multivariate} in \algline~\ref{free_gmm:hell_dist}.
To maintain mapping accuracy, we scale the distance threshold $\alpha_{h, \text{free}}$ using the geometric similarity $s_r \in [0, 1]$ between Gaussians $\gau{q_{g}}$ and $\gau{c_g}$ in \algline~\ref{free_gmm:sim_const}.
The geometric similarity $s_r$ between the two components is computed as the intersection over union ratio for the $z$ dimension of their bounding boxes in \algline~\ref{free_gmm:geo_sim}.

Even though free Gaussians are constructed to separately represent each subregion $\dist{B}_i$, the fusion decision made between Gaussians (\algline~\ref{free_gmm:sim_const}) in the current subregion $\dist{B}_i$ can be propagated to reduce the number of computations in subsequent subregions. 
The Gaussians recovered by the same basis across most subregions are almost relatively similar in shape (\emph{e.g.}, Gaussians $\gau{b}_{2,1}$, $\gau{b}_{1,1}$ and $\gau{b}_{0,1}$ in \fig~\ref{fig:free_space_fusion}).
Thus, the successful fusion between two Gaussians in the current subregion $\dist{B}_i$ (\emph{e.g.}, between $\gau{b}_{2,1}$ and $\gau{b}_{2,2}$) implies the same for other subregions $\dist{B}_{i-1}, \dots, \dist{B}_0$ (\emph{e.g.}, between $\gau{b}_{1,1}$ and $\gau{b}_{1,2}$, $\gau{b}_{0,1}$ and $\gau{b}_{0,2}$).
To automatically propagate the fusion decision from the current subregion $\dist{B}_i$, the fused basis $\gau{q_{f}}$ is simply transferred into the following subregions ($\dist{B}_{i-1}, \dots, \dist{B}_0$) at \algline~\ref{free_gmm:transfer_subregion} across multiple iterations of the outer loop (\algline~\ref{free_gmm:creation_start}).

\textbf{Construct Local Map} (\alglines~\ref{per_img:gmm_concat} to \ref{per_img:build_rtree} in \alg~\ref{alg:gmm_creation}): 
To enable the fusion between the local GMMs (occupied $\dist{G}_{t, \text{occ}}$ and free $\dist{G}_{t, \text{free}}$) and the global map $\dist{M}_{t-1}$ in Section~\ref{subsec:global_gmm_fusion}, local GMMs need to transform into the world frame as follows:
\begin{equation}\label{eqn:gau_transform}
	\mu_{\rv{X}} \leftarrow R_t\mu_{\rv{X}} + \epsilon_t,  \hspace{1ex}\Sigma_{\rv{X}} \leftarrow R_t \Sigma_{\rv{X}} R^\top_t,
\end{equation}
where $R_t$ and $\epsilon_t$ are the rotation and translation matrix associated with pose $T_t$. The mean $\mu_{\rv{X}}$ and covariance $\Sigma_{\rv{X}}$ for each Gaussian in the GMM are defined in \eqn~\eqref{eqn:gau_params}.

After the transformation, an R-tree is created for all Gaussians in \algline~\ref{per_img:build_rtree} to form the local map $\dist{G}_t$.
First, a bounding box is constructed for each Gaussian to enclose its ellipsoidal bound as defined in \eqn~\eqref{eqn:maha_dist}. Then, each Gaussian and its bounding box are inserted into the R-tree as shown in \fig~\ref{fig:gmm_creation}.

\subsection{Globally-Consistent GMM Fusion}~\label{subsec:global_gmm_fusion}

\begin{figure*}
	\centering
	\includegraphics[width=0.9\linewidth]{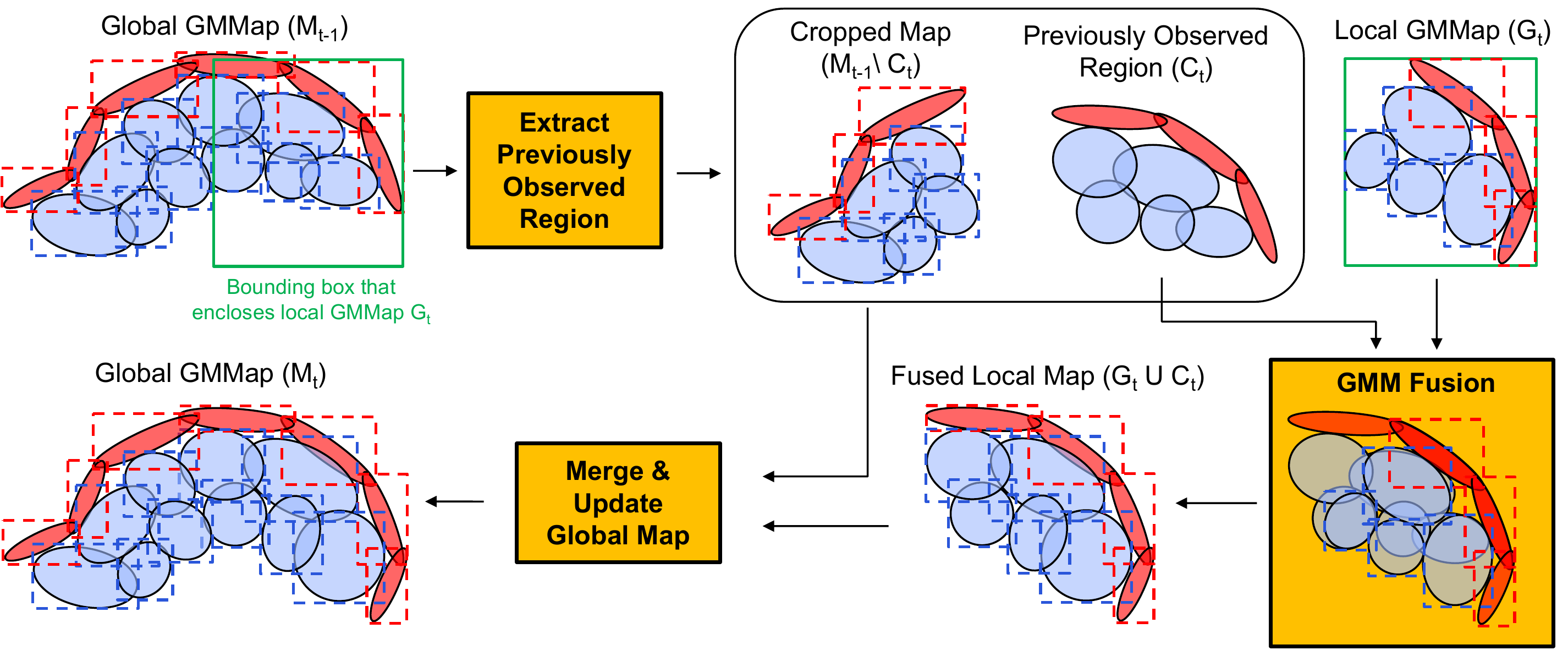}
	\caption{\textbf{Globally-consistent GMM fusion}: Constructing the current global GMMap $\dist{M}_t$ by fusing the local GMMap $\dist{G}_t$ into the previous global GMMap $\dist{M}_{t\text{--}1}$.  The bounding box (green rectangle) of local map $\dist{G}_t$ is used to determine the Gaussians $\dist{C}_t$ in the global map $\dist{M}_{t\text{--}1}$ that overlaps with $\dist{G}_t$ . Occupied and free GMMs are illustrated with red and blue ellipsoids, respectively. Dotted rectangles represent the bounding boxes at the leaf nodes of the R-tree. }
	\label{fig:gmm_fusion}
\end{figure*}

\begin{algorithm}[t!]\small
	\caption{Globally-Consistent GMM Fusion}\label{alg:gmm_fusion}
	\DontPrintSemicolon
	\KwIn{Local GMMap $\dist{G}_t$, previous global GMMap $\dist{M}_{t\text{--}1}$
	}
	\KwOut{Updated global GMMap $\dist{M}_t$}
	\SetKwBlock{Begin}{function}{end function}
	\Begin($\text{updateGlobalMap} {(} \dist{M}_{t\text{--}1}, \dist{G}_t {)}$)
	{
		$\dist{C}_t \leftarrow \text{findObservedRegion}(\dist{G}_t, \dist{M}_{t\text{--}1})$ \label{gmm_fusion:extract_ct}\;
		$\dist{M}_{t\text{--}1} \leftarrow \dist{M}_{t\text{--}1} \setminus \dist{C}_t$ \label{gmm_fusion:cropped_map}\;
		$\dist{M}_{t\text{--}1} \leftarrow \text{updateRtree} (\dist{M}_{t\text{--}1})$ \;
		\ForEach{$\gau{c} \in \dist{C}_t$}{\label{gmm_fusion:fusion_start}
			\eIf{$\textup{isFreeGaussian}(\gau{c})$}{
				$\dist{Q} \leftarrow \text{findIntersectingFreeGaussians}(\dist{G}_t, \gau{c})$\;
				$\alpha_h \leftarrow \alpha_{h, \text{free}}$ \;
			}{
				$\dist{Q} \leftarrow \text{findIntersectingOccGaussians}(\dist{G}_t, \gau{c})$\;
				$\alpha_h \leftarrow \alpha_{h, \text{occ}}$ \;
			}
			$\text{isObserved} \leftarrow \text{false}$ \;
			\ForEach{$\gau{q} \in \dist{Q}$}{
				$\gau{r} \leftarrow \text{fuseGaussians}(\gau{c}, \gau{q})$ \tcp*{See \eqn~\eqref{eqn:fuse_gau_gau}} \label{gmm_fusion:fuse_components}
				$d_h \leftarrow \text{unscentHellingerDistance}(\gau{r}, \gau{c}, \gau{q})$ \label{gmm_fusion:hell_dist}\;
				$s_r \leftarrow \text{geometricSimilarity}(\gau{c}, \gau{q})$ \label{gmm_fusion:geo_sim}\;
				\If{$d_h \leq s_r \cdot \alpha_h$}{\label{gmm_fusion:sim_const}
					$\gau{c} \leftarrow \gau{r}$ \label{gmm_fusion:gau_fusion}\;
					$\dist{G}_t \leftarrow \dist{G}_t \setminus \gau{q}$ \label{gmm_fusion:neighbor_remove}\;
					$\dist{G}_t \leftarrow \text{updateRtree} (\dist{G}_t)$ \;
					$\text{isObserved} \leftarrow \text{true}$ \;
				}
			}
			\If{$\textup{isObserved}$}{
				$\dist{C}_t \leftarrow \dist{C}_t \setminus \gau{c}$ \label{gmm_fusion:observed_remove}\;
				$\dist{G}_t \leftarrow \dist{G}_t \cup \gau{c}$ \label{gmm_fusion:append_local_map}\;
				$\dist{G}_t \leftarrow \text{updateRtree} (\dist{G}_t)$ \;
			}
		}\label{gmm_fusion:fusion_end}
		$\dist{M}_{t} \leftarrow \dist{M}_{t\text{--}1} \cup \dist{C}_t \cup \dist{G}_t$ \label{gmm_fusion:append_global}\;
		$\dist{M}_{t} \leftarrow \text{updateRtree} (\dist{M}_t)$ \;
		\Return{$\dist{M}_t$}
	}
\end{algorithm}

In this section, we present a novel memory-efficient procedure in \alg~\ref{alg:gmm_fusion} to directly update the global GMMap $\dist{M}_{t-1}$ in place using Gaussians from the local GMMap $\dist{G}_t$.
When the robot obtains a new depth image $Z_t$, rays associated with a subset of pixels in the image traverse through a previously observed region $\dist{C}_t$ that is modeled in the global map $\dist{M}_{t-1}$. 
To retain the compactness of the map, these rays should be fused with the map to update the state of the region $\dist{C}_t$.

In prior works~\cite{hornung2013octomap, Doherty2019, saarinen20133d, srivastava2018efficient}, the ray associated with each pixel in the image is cast into the global map to determine and update the region $\dist{C}_t$. 
Since these rays emanate outwards from the sensor origin, accessing the global map in memory along these rays often lacks spatial and temporal locality for effective cache usage.
Thus, casting all rays (more than 300,000 in each 640$\times$480 depth image) requires significant time and energy for accessing the DRAM where the map is stored.
In GMMap, the rays from the depth image are accurately compressed into a local GMMap $\dist{G}_t$ whose geometric properties are exploited to \emph{i)} quickly find the region $\dist{C}_t$ in the global map $\dist{M}_{t-1}$ with little memory access, and \emph{ii)} directly update the region $\dist{C}_t$ using Gaussians in $\dist{G}_t$ to maintain the compactness and accuracy of the resulting global map $\dist{M}_t$.

\fig~\ref{fig:gmm_fusion} illustrates the entire procedure for fusing the local map $\dist{G}_t$ into the global map $\dist{M}_{t-1}$.
Recall that from Section~\ref{subsec:per_img_gmm}, Gaussians in the local map $\dist{G}_t$ are already transformed in the world frame and organized using an R-tree. 
Using the bounding box (at the root node of the R-tree) that encloses $\dist{G}_t$, Gaussians in the previously observed region $\dist{C}_t$ can be extracted from the previous global map $\dist{M}_{t-1}$ in a \emph{single traversal} through its R-tree (\algline~\ref{gmm_fusion:extract_ct}) without ray casting.
After extraction, Gaussians in the region $\dist{C}_t$ are \emph{directly} fused with the local map $\dist{G}_t$ (\alglines~\ref{gmm_fusion:fusion_start} to \ref{gmm_fusion:fusion_end}). 
Since the Gaussians in $\dist{C}_t$ and $\dist{G}_t$ are extremely compact for storage within the on-chip cache, the entire fusion process is expected to require little DRAM accesses. 
After completion, the fused local map (\emph{i.e.}, $\dist{G}_t \cup \dist{C}_t$) is simply appended to the previous global map $\dist{M}_{t-1}$ in \algline~\ref{gmm_fusion:append_global} to produce the updated global map $\dist{M}_t$.

Our fusion process (\alglines~\ref{gmm_fusion:fusion_start} to \ref{gmm_fusion:fusion_end}) enhances the compactness of the local map $\dist{G}_t$ while maintaining its accuracy.
For each Gaussian $\gau{c} \in \dist{C}_t$, the R-tree in the local map $\dist{G}_t$ is used to efficiently search for the set of Gaussians $\dist{Q}$ that intersects with and represents the same type of region (\emph{i.e.}, free or occupied) as $\gau{c}$.
In \algline~\ref{gmm_fusion:fuse_components}, the Gaussian $\gau{c}$ is fused with each neighbor $\gau{q} \in \dist{Q}$ into a fusion candidate $\gau{r}$ using \eqn~\eqref{eqn:fuse_gau_gau}.
Similar to \alg~\ref{alg:free_gmm}, the fusion candidate $\gau{r}$ is accepted in \algline~\ref{gmm_fusion:gau_fusion} if the Unscented Hellinger distance~\cite{kristan2010multivariate} between candidate $\gau{r}$ and its components $\{\gau{c}, \gau{q}\}$ is less than a distance threshold $\alpha_h$.
Recall that our fusion criteria can exploit a wide range of scenarios (\emph{i.e.}, \fig~\ref{fig:overlap_fusion} and \ref{fig:neighbor_fusion}) to enhance the compactness of the map.
To maintain accuracy, we scale the distance threshold $\alpha_h$ in \algline~\ref{gmm_fusion:sim_const} using the geometric similarity $s_r$ (computed in \algline~\ref{gmm_fusion:geo_sim}) between components $\gau{c}$ and $\gau{q}$.
When $\gau{c}$ and $\gau{q}$ are free Gaussians that represent volumes, the similarity measure $s_r$ is the intersection over union ratio of their 3D bounding boxes. 
When $\gau{c}$ and $\gau{q}$ are occupied Gaussians that represent surfaces, the similarity measure $s_r$ is the intersection over union ratio of the largest two dimensions of their 3D bounding boxes (\emph{i.e.}, surface coverage similarity) multiplies by the dot product of their normal vectors (\emph{i.e.}, orientation similarity).
\section{Experimental Results \& Analysis}~\label{sec:results}
In this section, we compare our GMMap against current state-of-the-art frameworks \plcomment{in terms of accuracy, throughput, memory footprint, and energy consumption.}
Specifically, we chose the following frameworks with open-source implementations and different types of occupancy representations: OctoMap\footnote{[Online]. Available:  \url{https://github.com/OctoMap/octomap}}~\cite{hornung2013octomap} (discrete), NDT-OM\footnote{[Online]. Available:  \url{https://github.com/OrebroUniversity/perception_oru/tree/port-kinetic}}~\cite{saarinen20133d} (semi-parametric), and BGKOctoMap-L\footnote{[Online]. Available:  \url{https://github.com/RobustFieldAutonomyLab/la3dm}}~\cite{Doherty2019} (non-parametric).
This comparison was performed using four diverse indoor and outdoor environments (\emph{i.e.}, \emph{Room}, \emph{Warehouse}, \emph{Soulcity}, and \emph{Gascola}) generated from sequences of depth images and ground-truth poses.

\tab~\ref{tab:sequence_info} summarizes the characteristics of all four environments.
In particular, \emph{Room} (from real-world TUM-RGBD datasets~\cite{sturm12iros}) is a small structured environment that models crowded cubicles inside an office.
\emph{Warehouse} (from real-world TUM-RGBD datasets~\cite{sturm12iros}) is a larger structured indoor environment captured using a longer and noisier range of the Kinect camera.
In contrast, \emph{Soulcity} (from synthetic TartanAir dataset~\cite{wang2020tartanair}) is a large structured outdoor environment in a city containing several multi-story buildings with intricate sets of walkways.
Finally, \emph{Gascola} (from synthetic TartanAir dataset) is a large unstructured outdoor environment in a forest consisting of trees and a small hill.

To emulate an energy-constrained setting, all experiments were performed on the low-power NVIDIA Jetson TX2 platform in MAXP\_CORE\_ARM power mode~\cite{nvidia_tx2}.
All frameworks, implemented in C++, were compiled using the same settings.
To reduce the memory overhead and map size, the floating point variables (and their associated operations) across all frameworks are stored as (and performed in) 32-bit single precision. 
Our single-core, multi-core, and GPU-accelerated GMMap implementations (visualized in Open3D~\cite{open3d}) can be obtained at \href{https://github.com/mit-lean/GMMap}{\color{blue}{https://github.com/mit-lean/GMMap}}.

Prior works achieve high mapping accuracy but are neither as computationally nor memory efficient as GMMap.
After carefully selecting a set of hyperparameters for all frameworks (Section~\ref{subsec:hyperparameters}), GMMap is evaluated across a diverse set of indoor and outdoor environments and achieves comparable accuracy as prior works (Section~\ref{subsec:accuracy}).
In addition, our GMMap is highly parallelizable and can be constructed in real-time at up to 81 images per second, which is $4\times$ to $146\times$ higher than prior works on the low-power Jetson TX2 platform (Section~\ref{subsec:throughput}).
Due to single-pass depth image compression in SPGF* and directly operating on Gaussians during map construction, our GMMap is extremely memory efficient. Compared with prior works (in Section~\ref{subsec:memory}), our CPU implementation reduces \emph{i)} the map size by at least 56\%, \emph{ii)} the memory overhead for storing input and temporary variables by at least 88\%, and \emph{iii)} the number of DRAM accesses by at least 78\% during map construction.
Thus, in Section~\ref{subsec:energy_consumption}, the computational and memory efficiency of our GMMap reduces energy consumption by at least 69\% compared with prior works.

\begin{table}[!t]
	\centering
	\caption{Properties of all four environments used for evaluation}
	\resizebox{\columnwidth}{!}{
		\begin{tabular}{ccccc} 
			\toprule
			\textbf{Environment} &
			\makecell{\textbf{Dimensions (m)}} & \textbf{Images} & \makecell{\textbf{Depth Image} \\ \textbf{Resolution}} & \makecell{\textbf{Avg. Sensor} \\ \textbf{Range (m)}} \\ 
			\midrule 
			\makecell{Room \\ (freiburg1\_room)} & 11.28 $\times$ 12.05 $\times$ 3.45 & 1311 & 640$\times$480 & 0.97 \\
			\midrule
			\makecell{Warehouse \\ (freiburg2\_ \\ pioneer\_slam)} & 23.52 $\times$ 17.90 $\times$ 4.29 & 2169 & 640$\times$480 & 1.13 \\
			\midrule
			Soulcity & 73.90 $\times$ 62.41 $\times$ 42.69 & 1083 & 640$\times$480 & 10.85 \\
			\midrule
			Gascola & 59.04 $\times$ 52.93 $\times$ 33.71 & 382 & 640$\times$480 & 4.06 \\
			\bottomrule
		\end{tabular}
		\label{tab:sequence_info}
	}
\end{table}

\subsection{Selection of Hyperparameters}~\label{subsec:hyperparameters}
In this section, we discuss the selection of hyperparameters for the GMMap (Section~\ref{subsubsec:hyper_gmmap}). Then, we briefly summarize the hyperparameters selected for prior frameworks (\emph{i.e.}, NDT-OM, BGKOctoMap-L, and OctoMap, in Section~\ref{subsubsec:hyper_existing}).  
The hyperparameters of all frameworks are presented in \tab~\ref{tab:map_param} and are manually tuned to reduce the size of the maps without significant deviation from their peak accuracy.

\subsubsection{\textbf{GMMap}}~\label{subsubsec:hyper_gmmap}
For GMMap, the hyperparameters are the unexplored prior weight $\pi_0$ for the prior distribution in \eqn~\eqref{eqn:prior_dist}, the initial distance $d_0$ between partitioning planes in \eqn~\eqref{eqn:partition_plane}, and the distance thresholds ($\alpha_{h, \text{free}}$ and $\alpha_{h, \text{occ}}$) for fusing Gaussians in \alg~\ref{alg:free_gmm} and \ref{alg:gmm_fusion}. 
\plcomment{These hyperparameters are mostly dependent on the sensor itself (\emph{i.e.}, not on the environment). Recall that the Gaussian fusion thresholds ($\alpha_{h, \text{free}}$ and $\alpha_{h, \text{occ}}$) control the trade-off between compactness and the accuracy of the map.
	For environments (\emph{i.e.}, \emph{Room} and \emph{Warehouse}) captured using a noisy sensor (\emph{i.e.}, Kinect), we set both thresholds lower to better capture the surface details and the noise of the sensors at the expense of compactness, as shown in \tab~\ref{tab:map_param}.
	For environments (\emph{i.e.}, \emph{Soulcity} and \emph{Gascola}) captured using an ideal noiseless sensor,  Gaussians that represent the same region appear much more similar. Thus, we set both fusion thresholds higher to achieve better compactness without sacrificing accuracy.
	
	Recall that from \eqn~\eqref{eqn:partition_plane}, the (initial) partition plane distance $d_0$ affects the locations and number of free space subregions in each local map. Since the field of views for our cameras are quite similar across all environments, the distance $d_0$ is almost constant in \tab~\ref{tab:map_param}.
	Finally, the unexplored prior weight $\pi_0$ controls the evidence of the unexplored region. Thus, a larger $\pi_0$ would require \emph{i)} more measurements for the occupancy probability to converge from unexplored state (\emph{i.e.}, 0.5) to free (\emph{i.e.}, 0) or occupied state (\emph{i.e.}, 1), and \emph{ii)} reduce the amount of free region spatially interpolated from the frontier (\emph{i.e.}, the boundary between free and unexplored region) into the unexplored region.
	Since the number of pixels in the depth image is constant across all environments, the unexplored prior weight $\pi_0$ is unchanged in \tab~\ref{tab:map_param} to retain the same convergence rate of occupancy across environments.}

\begin{table*}[!t]
	\centering
	\caption{Hyperparameters used in GMMap, NDT-OM, BGKOctoMap-L, and OctoMap across all four environments}
	\resizebox{\textwidth}{!}{
		\begin{tabular}{c|cccc|c|ccc|c} 
			\toprule
			\multirow{2}{*}{\makecell{\\ \textbf{Environment} \\ }} & \multicolumn{4}{c|}{\textbf{GMMap}} & \textbf{NDT-OM} & \multicolumn{3}{c|}{\textbf{BGKOctoMap-L}} & \textbf{OctoMap} \\
			& \makecell{Unexplored Prior \\ Weight ($\pi_0$)} & \makecell{Partition Plane \\ Distance ($d_0$)} & \makecell{Free Gaussian \\ Fusion Threshold ($\alpha_{h, \text{free}}$)} & \makecell{Occupied Gaussian \\ Fusion Threshold ($\alpha_{h, \text{occ}}$)} & \makecell{Voxel \\ Size} & \makecell{Voxel \\ Size} & \makecell{Free \\ Resolution} & \makecell{Block Octree \\ Depth} & \makecell{Voxel \\ Size} \\
			\midrule    
			Room & 500,000 & 0.5m & 0.26 & 0.70 & 0.4m & 0.1m & 0.3m & 3 & 0.1m \\
			Warehouse & 500,000 & 0.5m & 0.26 & 0.70 & 0.5m & 0.1m & 0.3m & 3 & 0.2m \\
			Soulcity & 500,000 & 0.5m & 0.63 & 1.41 & 1.2m & 0.3m & 3.0m & 3 & 0.3m \\
			Gascola & 500,000 & 0.6m & 0.63 & 1.41 & 1.2m & 0.3m & 3.0m & 3 & 0.3m \\
			\bottomrule
		\end{tabular}
		\label{tab:map_param}
	}
\end{table*}

\subsubsection{\textbf{Existing Frameworks}}~\label{subsubsec:hyper_existing}
For NDT-OM, BGKOctoMap-L, and OctoMap, the environment is voxelized so that the minimum voxel size is the hyperparameter. \plcomment{In our experiments, we increase the voxel size with the size of the environment to maintain a good trade-off between accuracy and the compactness of the map in \tab~\ref{tab:map_param}.}
Since BGKOctoMap-L partitions the environment into equally-sized cubic blocks such that each block contains an octree (\emph{i.e.}, defined as the test-data octrees~\cite{Doherty2019}), the depth of the octree in each block is an additional hyperparameter which is kept constant across all environments.
To generate compact training data representing free regions, BGKOctoMap-L samples points along each sensor ray at a free resolution interval. \plcomment{In \tab~\ref{tab:map_param}, we increase this interval with the size of the environment to maintain a good trade-off between accuracy and map construction throughput.}

\subsection{Accuracy of Occupancy Estimation}~\label{subsec:accuracy}
\begin{figure*}
	\centering
	
	\subfloat[GMMap (Gaussians)]{%
		\includegraphics[width=0.18\textwidth]{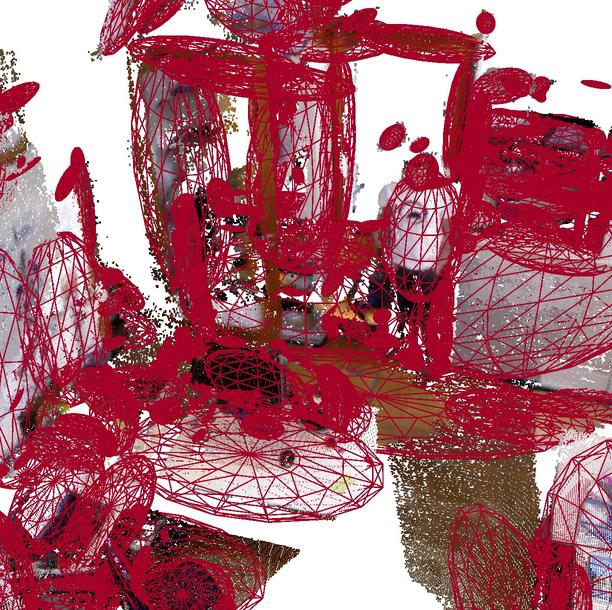}
		\label{fig:gmmap_room}
	}
	\hfill
	\subfloat[\plcomment{GMMap (Voxels)}]{%
		\includegraphics[width=0.18\textwidth]{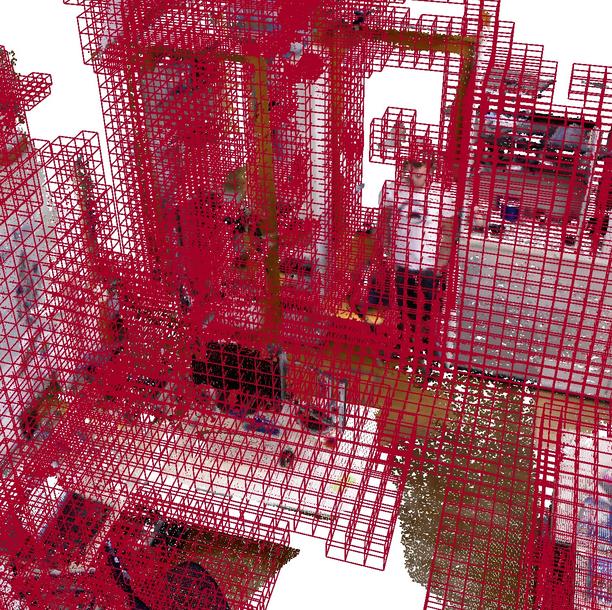}
		\label{fig:gmmap_room_voxel}
	}
	\hfill
	\subfloat[NDT-OM]{%
		\includegraphics[width=0.18\textwidth]{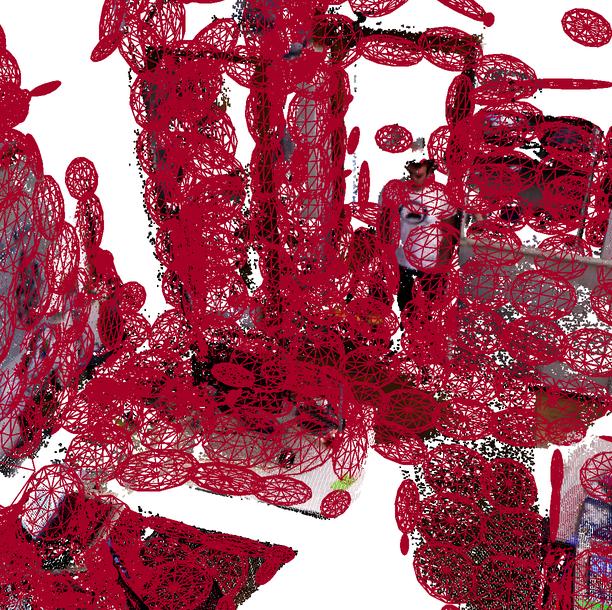}
		\label{fig:ndt_room}
	}
	\hfill
	\subfloat[BGKOctoMap-L]{%
		\includegraphics[width=0.18\textwidth]{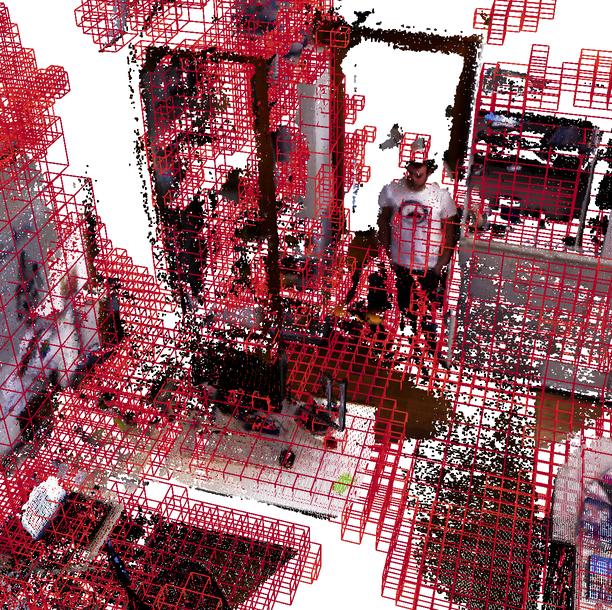}
		\label{fig:bgkoctomapl_room}
	}
	\hfill
	\subfloat[OctoMap]{%
		\includegraphics[width=0.18\textwidth]{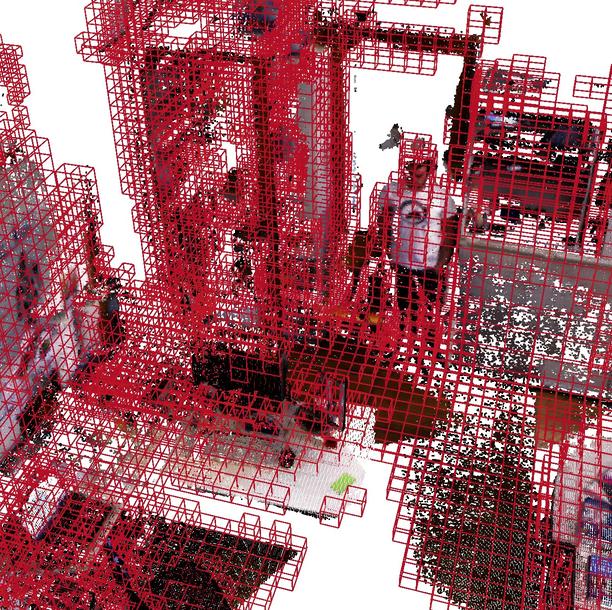}
		\label{fig:octomap_room}
	}
	
	\caption{Visualization of the ground-truth \emph{Room} (structured indoor) environment overlaid with the following mapping frameworks: (a) GMMap (occupied Gaussians), (b) GMMap (occupied voxels from uniform voxel-grid sampling), (c) NDT-OM (occupied Gaussians), (d) BGKOctoMap-L (occupied voxels), and (e) OctoMap (occupied voxels).
	Free regions are not illustrated for ease of visualization.
	\plcomment{Even though Gaussians are unbounded and continuous, each occupied Gaussian is visualized at a Mahalanobis distance of two using an ellipsoidal wireframe in (a) and (c). For BGKOctoMap-L, OctoMap, and GMMap (from uniform voxel-grid sampling), wireframes of occupied voxels with an occupancy probability greater than 0.9 are visualized. For both GMMap and NDT-OM, the surface boundaries of the obstacles are smooth and often \emph{extend beyond their ellipsoidal wireframes} (\emph{e.g.}, compare (a) and (b) for GMMap).}}
	\label{fig:room_illustrations} 
\end{figure*}

\begin{figure*}
	\centering
	\subfloat[GMMap (Gaussians)]{%
		\includegraphics[width=0.18\textwidth]{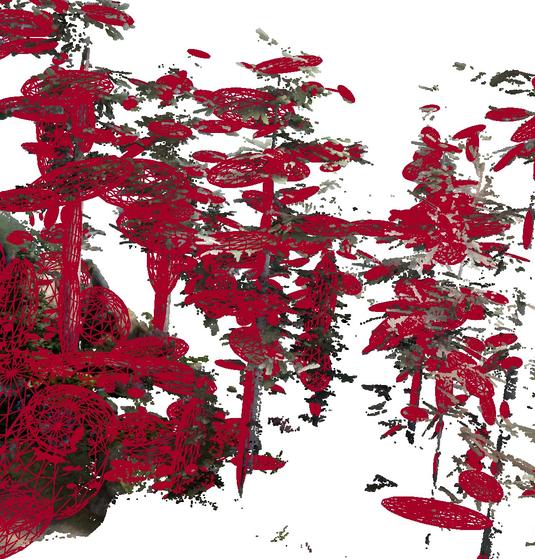}
		\label{fig:gmmap_gascola}
	}
	\hfill
	\subfloat[\plcomment{GMMap (Voxels)}]{%
		\includegraphics[width=0.18\textwidth]{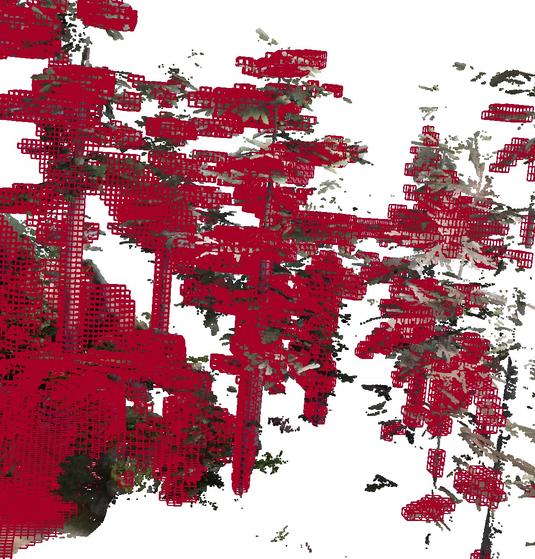}
		\label{fig:gmmap_gascola_voxel}
	}
	\hfill
	\subfloat[NDT-OM]{%
		\includegraphics[width=0.18\textwidth]{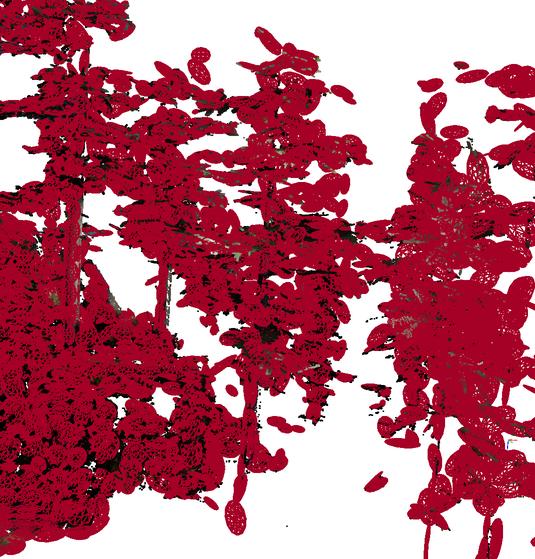}
		\label{fig:ndt_gascola}
	}
	\hfill
	\subfloat[BGKOctoMap-L]{%
		\includegraphics[width=0.18\textwidth]{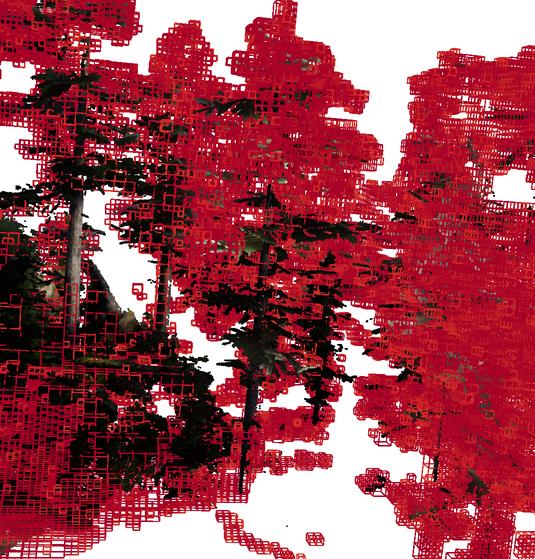}
		\label{fig:bgkoctomapl_gascola}
	}
	\hfill
	\subfloat[OctoMap]{%
		\includegraphics[width=0.18\textwidth]{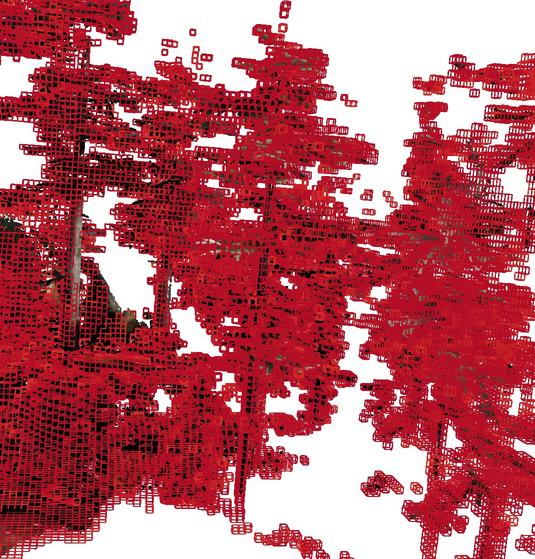}
		\label{fig:octomap_gascola}
	}
	
	\caption{Visualization of the ground-truth \emph{Gascola} (unstructured outdoor) environment overlaid with the following mapping frameworks: (a) GMMap (occupied Gaussians), (b) GMMap (occupied voxels from uniform voxel-grid sampling), (c) NDT-OM (occupied Gaussians), (d) BGKOctoMap-L (occupied voxels), and (e) OctoMap (occupied voxels). 
	Free regions are not illustrated for ease of visualization.	
	\plcomment{Even though Gaussians are unbounded and continuous, each occupied Gaussian is visualized at a Mahalanobis distance of two using an ellipsoidal wireframe in (a) and (c). For BGKOctoMap-L, OctoMap, and GMMap (from uniform voxel-grid sampling), wireframes of occupied voxels with an occupancy probability greater than 0.9 are visualized. For both GMMap and NDT-OM, the surface boundaries of the obstacles are smooth and often \emph{extend beyond their ellipsoidal wireframes} (\emph{e.g.}, compare (a) and (b) for GMMap).}}
	\label{fig:gascola_illustrations} 
\end{figure*}

In this section, we compare the accuracy of the GMMap against NDT-OM, BGKOctoMap-L, and OctoMap.
\plcomment{Specifically, we are interested in the accuracy of occupancy estimation in the occupied and free regions (Section~\ref{subsubsec:acc_occ_free_region}) as well as the characteristics of the obstacle surfaces (at the free-to-occupied regions, in Section~\ref{subsubsec:acc_obs_surface}) and frontiers (at the free-to-unexplored regions, in Section~\ref{subsubsec:acc_frontier}).}
Recall that the hyperparameters in \tab~\ref{tab:map_param} are manually tuned to reduce the size of the maps without significant deviation from their peak accuracy for estimating occupancy probability. 
\plcomment{Depending on the downstream applications, these hyperparameters are also used to control two fundamental trade-offs within the GMMap, as discussed in Sections~\ref{subsubsec:acc_occ_free_region} and \ref{subsubsec:acc_frontier}.}
\plcomment{For our experiments, we evaluated the occupancy probabilities at the end of all sensor rays for validating occupied regions, and along all sensor rays at 10 cm intervals for validating free regions.}
\fig~\ref{fig:room_illustrations} and \ref{fig:gascola_illustrations} illustrate each framework in \emph{Room} (structured indoor) and \emph{Gascola} (unstructured outdoor) environment, respectively.

\subsubsection{\textbf{Occupied \& Free Regions}}~\label{subsubsec:acc_occ_free_region}
We use the receiver operating characteristics (ROC) curve to compare the accuracy for estimating occupancy in both occupied and free regions across all frameworks.
To generate the ROC curves, the occupancy probability is queried from each map at the locations of all sensor rays used to construct the map. 
%
%
By sweeping the thresholds for classifying occupied or free regions from each occupancy probability, the true positive rate (\emph{i.e.}, the proportion of correct classifications during the prediction of occupied regions) of each map varies with the false positive rate (\emph{i.e.}, the proportion of incorrect classifications during the prediction of the free regions). 
In addition, the area under the curve (AUC) represents the probability that the map estimates a higher occupancy for the occupied region than that of the free region~\cite{fawcett2006introduction}.
Thus, a map with high accuracy should generate a ROC curve that tends towards the upper-left corner of the plot to achieve a large AUC close to one.

\begin{figure*}
	\centering
	\subfloat[\emph{Room}\label{roc_a}]{%
		\includegraphics[width=0.23\textwidth]{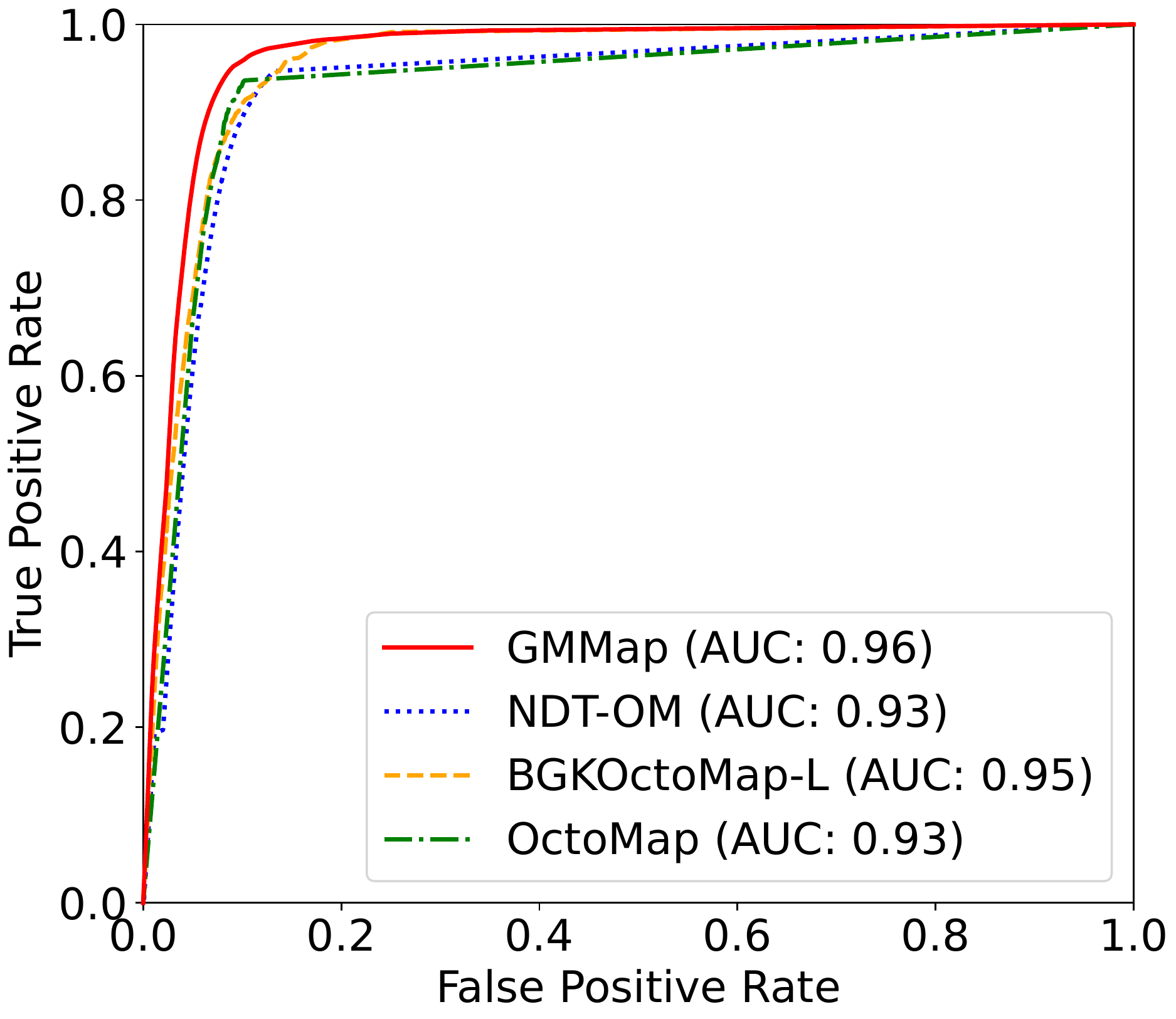}
		\label{fig:roc_room}
	}
	\hfill
	\subfloat[\emph{Warehouse}\label{roc_b}]{%
		\includegraphics[width=0.23\textwidth]{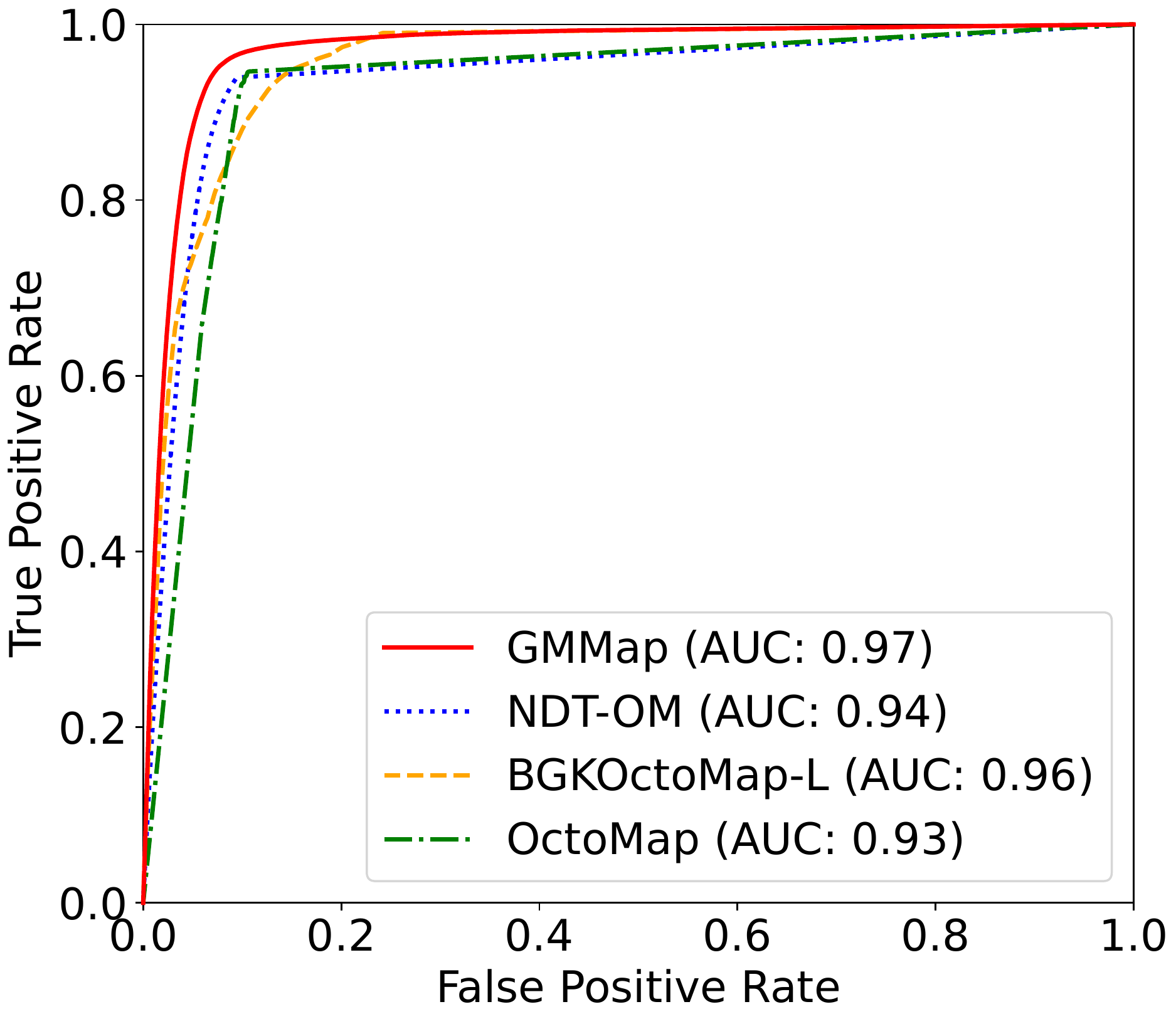}
		\label{fig:roc_warehouse}
	}
	\hfill
	\subfloat[\emph{Soulcity}\label{roc_c}]{%
		\includegraphics[width=0.23\textwidth]{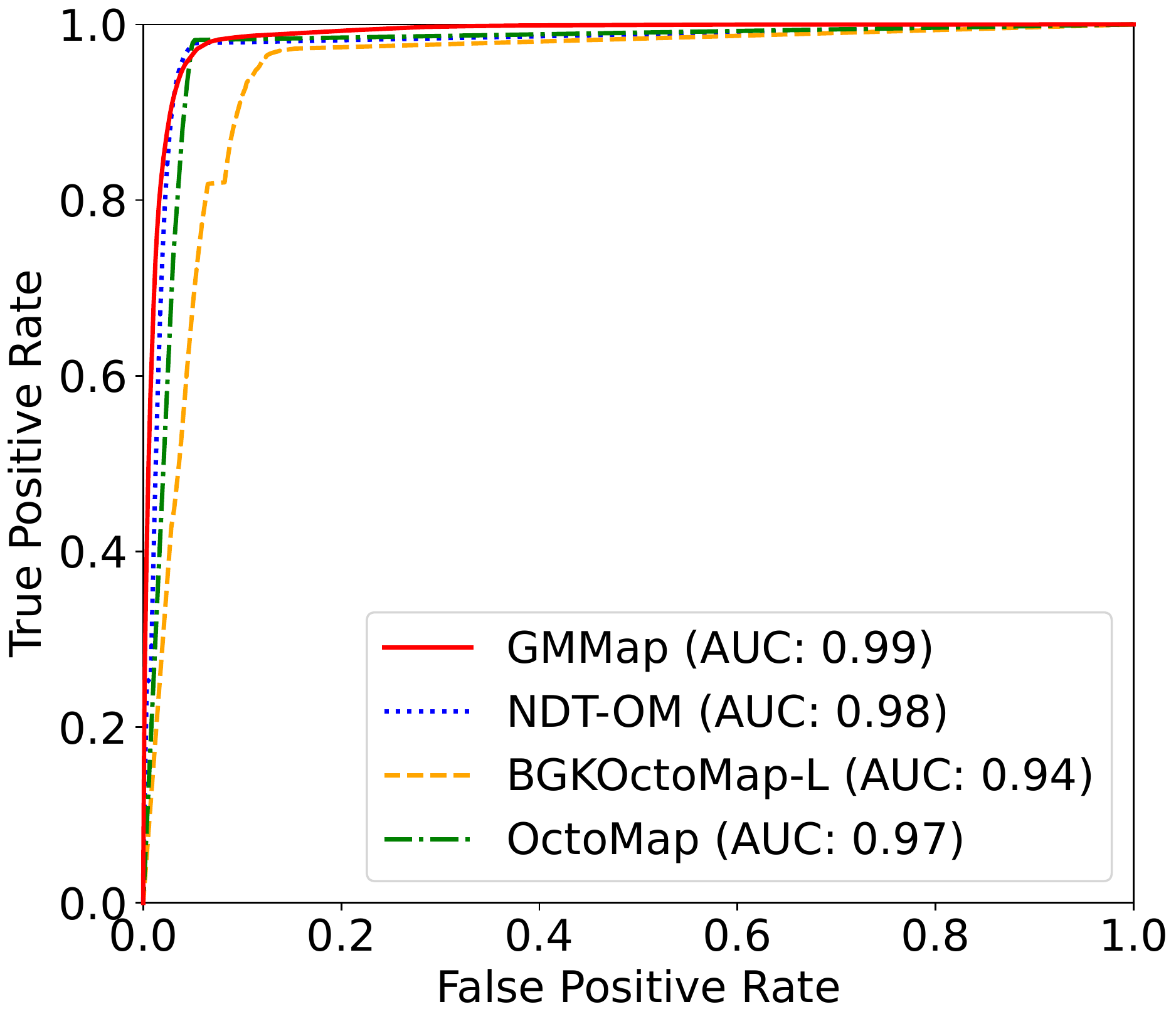}
		\label{fig:roc_soulcity}
	}
	\hfill
	\subfloat[\emph{Gascola}\label{roc_d}]{%
		\includegraphics[width=0.23\textwidth]{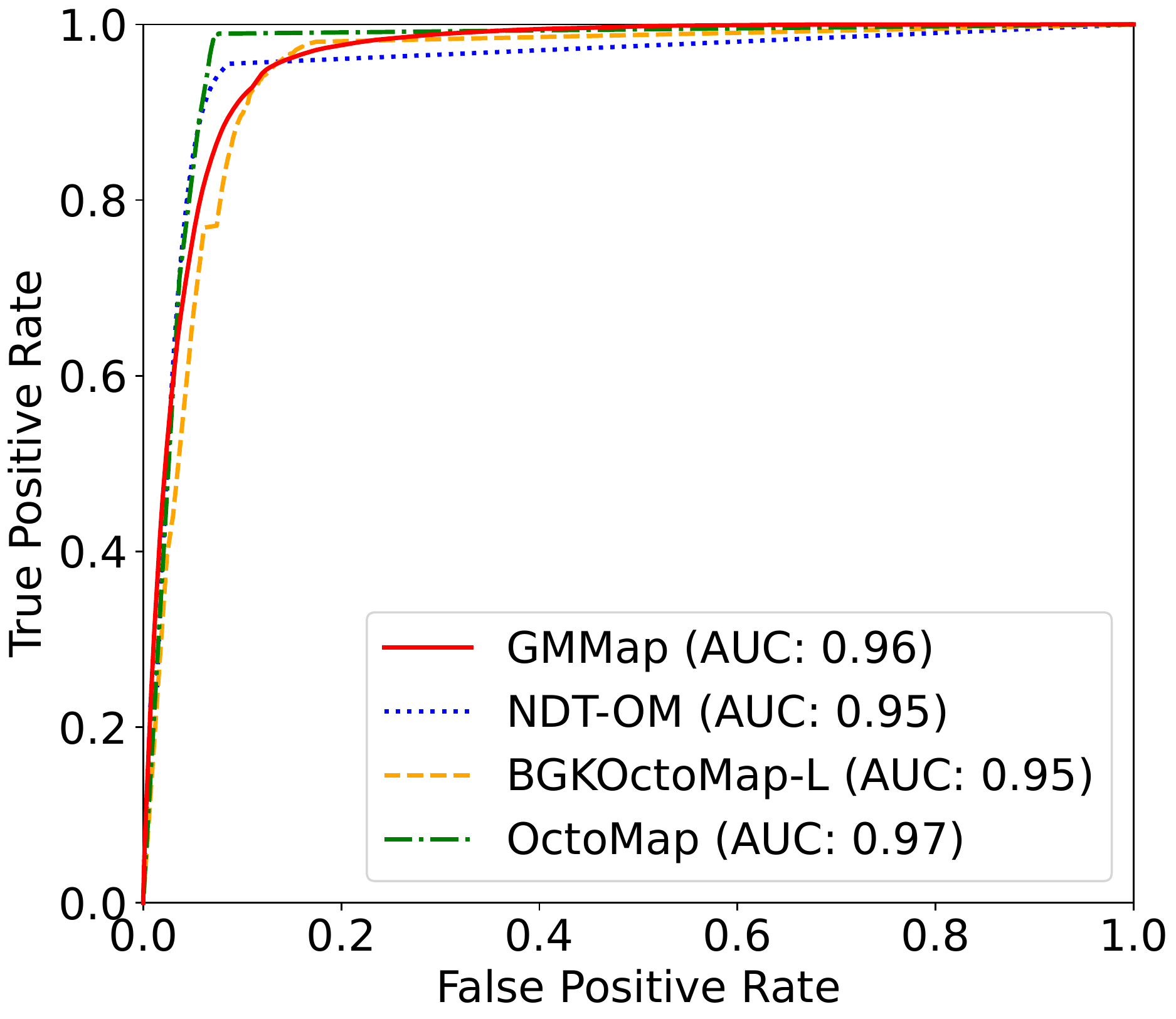}
		\label{fig:roc_gascola}
	}
	
	\caption{Comparison of receiver operating characteristic (ROC) for classifying occupied and free regions among GMMap, OctoMap, NDT-OM, and BGKOctoMap-L in four environments: (a) \emph{Room}, (b) \emph{Warehouse}, (c) \emph{Soulcity}, and (d) \emph{Gascola}. 
		The area under the ROC curve (AUC) equals to the probability that an occupied region is assigned a higher occupancy probability than the free region in the map.
		From the AUC of the ROC curves, the accuracy of occupancy estimation in occupied and free regions for the GMMap is comparable with other frameworks across all environments.}
	\label{fig:roc} 
\end{figure*}

\fig~\ref{fig:roc} illustrates the ROC curve for each framework across all environments. The (ROC) AUC for the GMMap is slightly higher than other frameworks in structured indoor (\emph{i.e.}, \emph{Room} and \emph{Warehouse}) and outdoor (\emph{i.e.}, \emph{Soulcity}) environments.
\plcomment{In addition, the ROC curves of other frameworks are mostly under the ROC curves of GMMap, which indicates Gaussians' ability to accurately model both volumetric free regions and thin obstacle surfaces with fewer false positives (\emph{i.e.}, mistakenly predicting free regions as occupied) without sacrificing true positives (\emph{i.e.}, correctly predicting occupied regions).
In contrast, voxels in OctoMap and BGKOctoMap-L inflate the obstacle surfaces with cubic artifacts whose thickness is at least equal to the minimum voxel size, which leads to higher false positives.}
Even though NDT-OM also utilizes Gaussians to represent obstacle surfaces, these Gaussians are constructed under the assumption that all measurements within each voxel belong to the same surface, and thus also suffer from voxelization artifacts when such assumption is invalid (\emph{e.g.}, at corners of objects in \fig~\ref{fig:ndt_room}).

For unstructured outdoor environment (\emph{i.e.}, \emph{Gascola}), the (ROC) AUC of GMMap is comparable but slightly lower than OctoMap as shown in \fig~\ref{fig:roc_gascola}.
To achieve a compact representation and avoid modeling spurious measurements, GMMap prunes away small occupied Gaussians containing less than a certain number of measurements (\emph{i.e.}, 200 in our experiments).
While this pruning threshold does not affect the modeling of objects closer to the camera (typically containing more measurements), this threshold prunes away occupied Gaussians for small objects (\emph{i.e.}, leaves on the tree in \fig~\ref{fig:gmmap_gascola}) that are far away from the camera (and robot).

\plcomment{
	In fact, the Gaussian pruning threshold controls one of the fundamental trade-offs between the ability to model small distant objects and the compactness of the GMMap.
	For applications that require the mapping of distant small objects, the pruning threshold can be decreased to retain these objects at the expense of a larger map size.
	However, pruning these small occupied Gaussians does not degrade the safety of the map.
	In SPGF*, both the occupied regions associated with these pruned Gaussians and free regions near them will remain unexplored because free Gaussian bases leading up to (and associated with) these occupied Gaussians are also pruned.
 	This is reflected in \fig~\ref{fig:roc_gascola} where the ROC curve of GMMap shifts to the right (\emph{i.e.}, higher false positives for free regions near these pruned Gaussians) compared with other frameworks.
	If the robot recaptures these previously-neglected small objects at a closer distance later, these objects will be modeled by more measurements and thus retained in GMMap.
}

\begin{figure*}
	\centering
	\subfloat[\plcomment{\emph{Room}}\label{pr_a}]{%
		\includegraphics[width=0.23\textwidth]{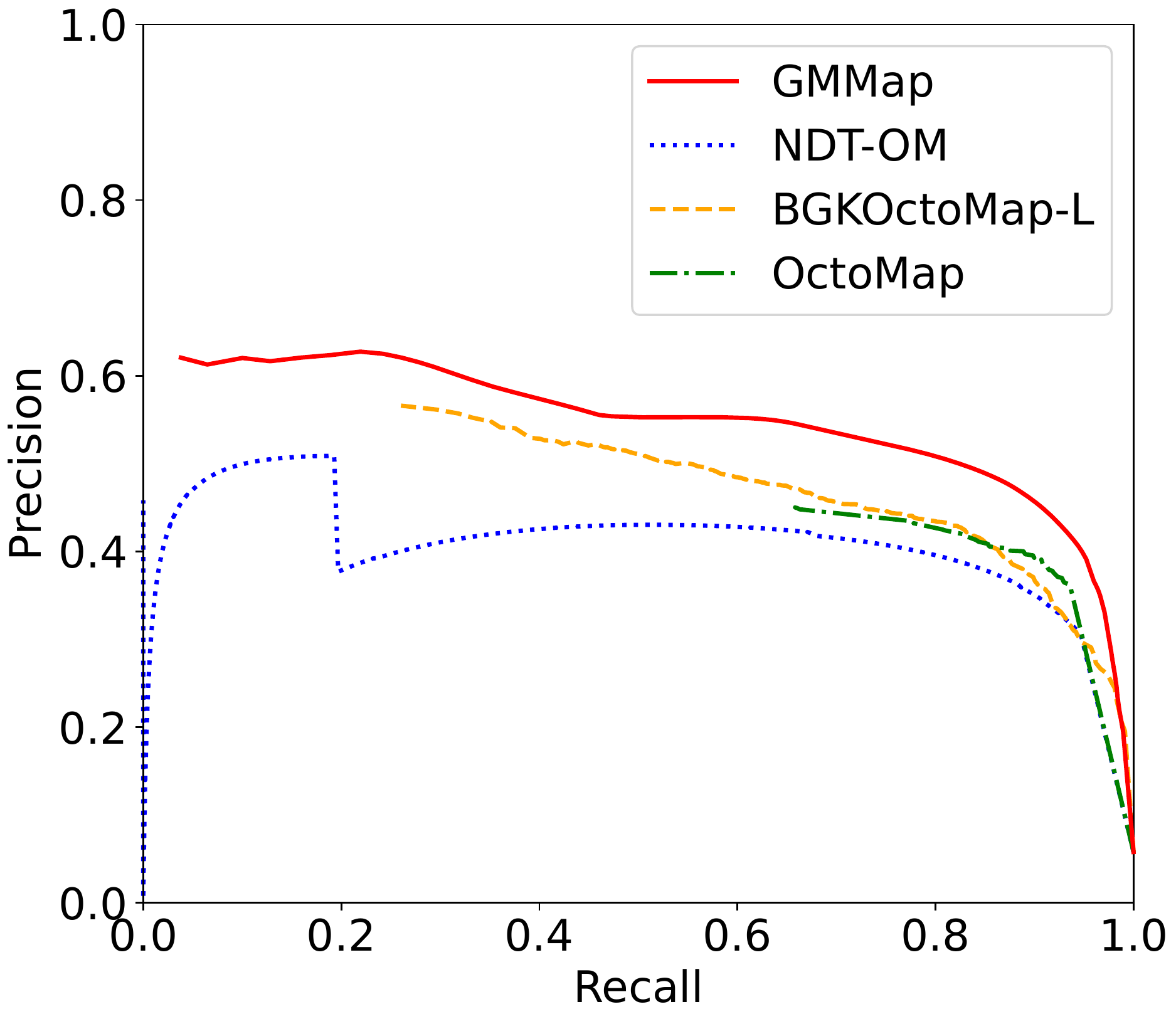}
		\label{fig:pr_room}
	}
	\hfill
	\subfloat[\plcomment{\emph{Warehouse}}\label{pr_b}]{%
		\includegraphics[width=0.23\textwidth]{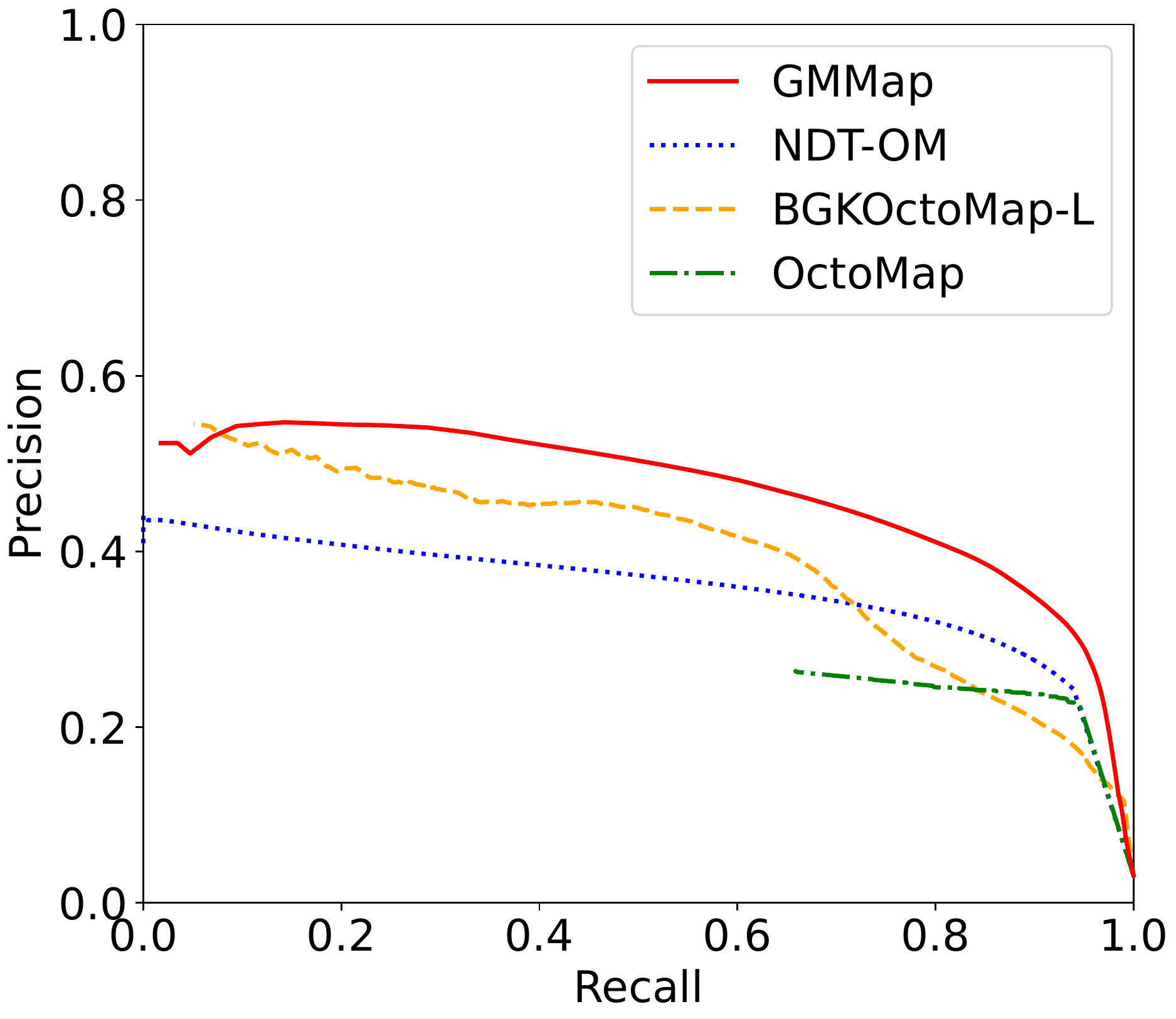}
		\label{fig:pr_warehouse}
	}
	\subfloat[\plcomment{\emph{Soulcity}}\label{pr_c}]{%
		\includegraphics[width=0.23\textwidth]{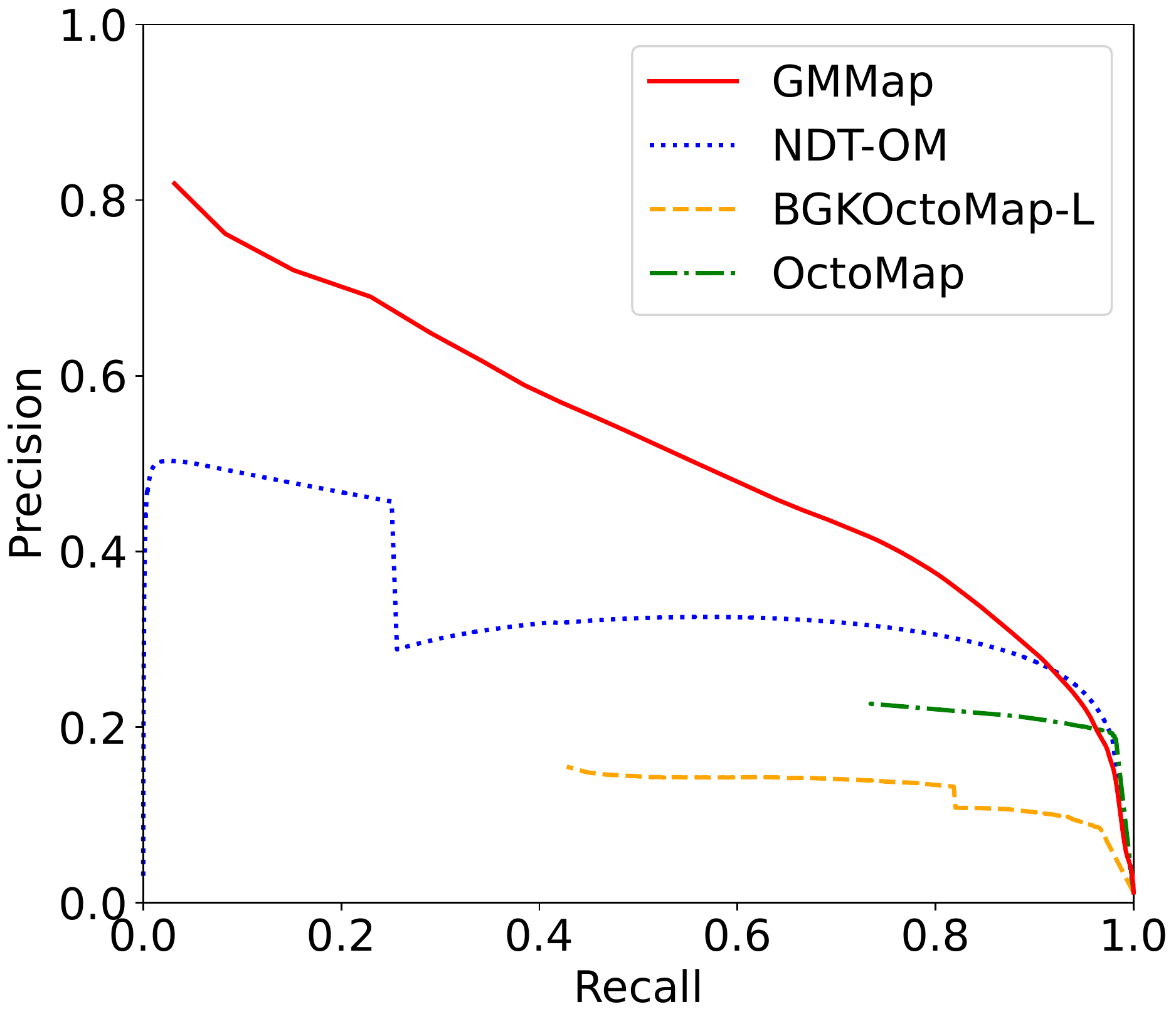}
		\label{fig:pr_soulcity}
	}
	\hfill
	\subfloat[\plcomment{\emph{Gascola}}\label{pr_d}]{%
		\includegraphics[width=0.23\textwidth]{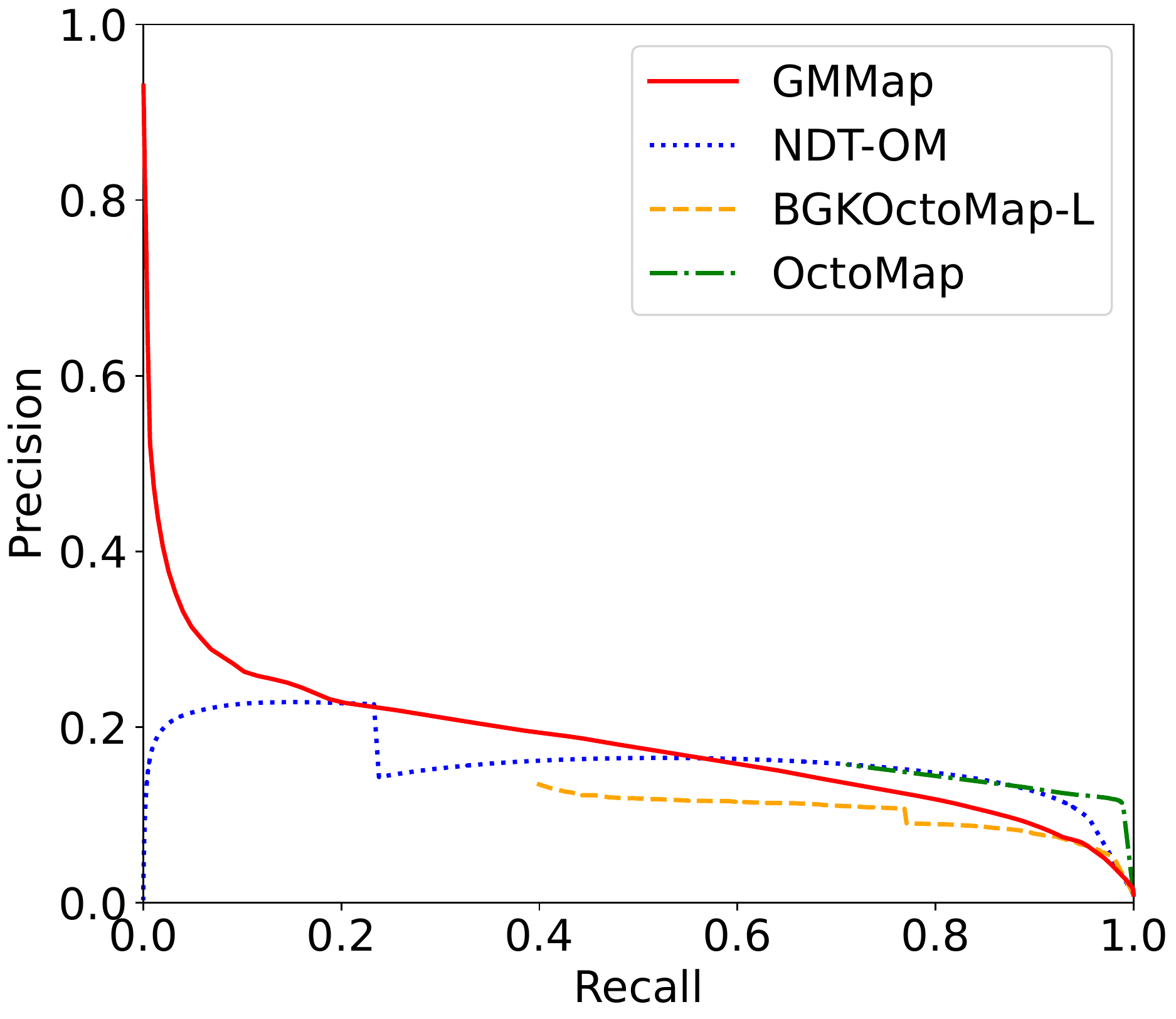}
		\label{fig:pr_gascola}
	}
	
	\caption{
		\plcomment{
			Comparison of precision-recall curves for modeling obstacle surfaces among GMMap, OctoMap, NDT-OM, and BGKOctoMap-L in four environments: (a) \emph{Room}, (b) \emph{Warehouse}, (c) \emph{Soulcity}, and (d) \emph{Gascola}.
			No precision-recall curve can reach the y-axis because precision becomes invalid when the occupancy decision threshold equals or exceeds the maximum occupancy probability in the map (\emph{i.e.}, 0.97 of OctoMap and 1.0 for others).
			In addition, no framework achieves 100\% precision due to the inflation of obstacle surfaces from voxelization (in OctoMap and BGKOctoMap-L) or smooth decay of Gaussians (in GMMap and NDT-OM).
			For NDT-OM and GMMap, the occupancy probability converges to one at the mean of each occupied Gaussian.
			Since NDT-OM models all obstacle surfaces within each voxel by a single Gaussian, the mean of such Gaussian could be located in a free region between multiple surfaces which leads to a sharp decrease in both precision and recall at high occupancy decision threshold.
			Since Gaussians in the GMMap overlap and flexibly adapt to the thickness, location, and orientation of each surface, GMMap achieves higher precision at similar recall than other frameworks across most environments.}}
	\label{fig:pr} 
\end{figure*}

\subsubsection{\textbf{Obstacle Surfaces (at Free-to-Occupied Regions)}}~\label{subsubsec:acc_obs_surface}
\plcomment{
We used the precision-recall curves to compare the modeling of the obstacle surfaces among all frameworks at different occupancy decision thresholds.
In \fig~\ref{fig:pr}, no precision-recall curve can reach the y-axis because precision becomes invalid when the occupancy decision threshold equals or exceeds the maximum occupancy probability in the map (\emph{i.e.}, 0.97 for OctoMap and 1.0 for others).
In addition, no framework achieves 100\% precision due to the inflation/thickening of obstacle surfaces from either voxelization (in OctoMap and BGKOctoMap-L) or smooth decay of Gaussians (in GMMap and NDT-OM). 
To ensure safe navigation, we are interested in the precision of obstacle surfaces when the recall is high.
In structured environments (\emph{i.e.}, \emph{Room}, \emph{Warehouse}, and \emph{Soulcity}), GMMap often achieves higher precision at similar recall compared with other frameworks because Gaussians in the GMMap overlap and flexibly adapt to the thickness, location, and orientation of each surface.
In other frameworks, the precision is lower due to higher false positives caused by \emph{i)} using Gaussian to model multiple distinct surfaces in each voxel (in NDT-OM) or \emph{ii)} thickening of surfaces induced by the minimum size of the voxels (in OctoMap and BGKOctoMap-L).
For unstructured outdoor environment (\emph{i.e.}, \emph{Gascola} in \fig~\ref{fig:pr_gascola}), the precision of GMMap is lower than other frameworks at higher recall due to the pruning of Gaussians representing small distant objects far away from the robot.
As explained in Section~\ref{subsubsec:acc_occ_free_region}, the locations of these pruned Gaussians will remain unexplored and thus do not affect the safety of navigation.
%
%
%
}

\subsubsection{\textbf{Frontiers (at Free-to-Unexplored Regions)}}~\label{subsubsec:acc_frontier}
\begin{figure*}
	\centering
%
	\subfloat[\emph{Warehouse} point cloud]{%
		\includegraphics[width=0.32\textwidth]{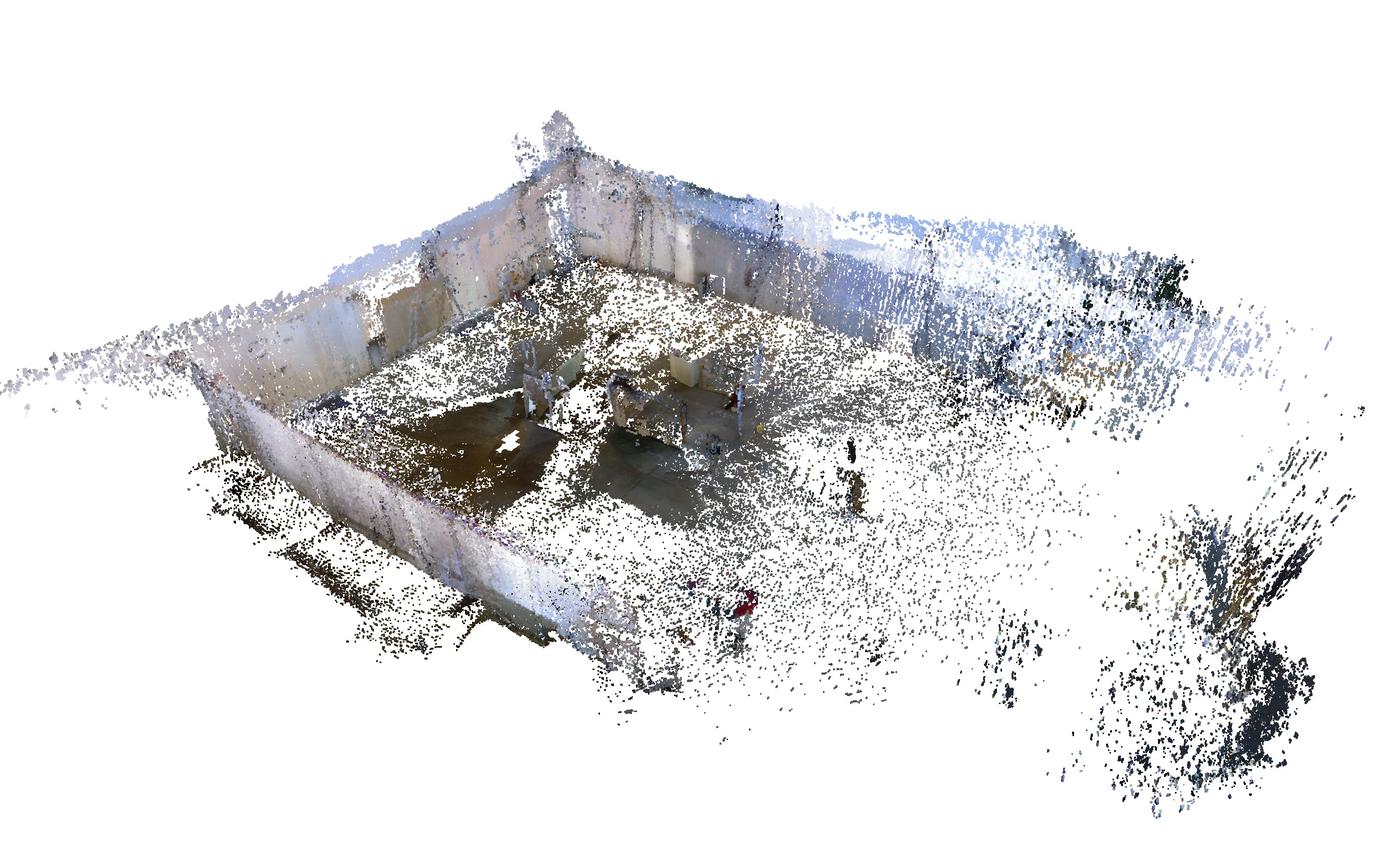}
		\label{fig:warehouse_pcd}
	}
	\hfill
	\subfloat[GMMap]{%
		\includegraphics[width=0.32\textwidth]{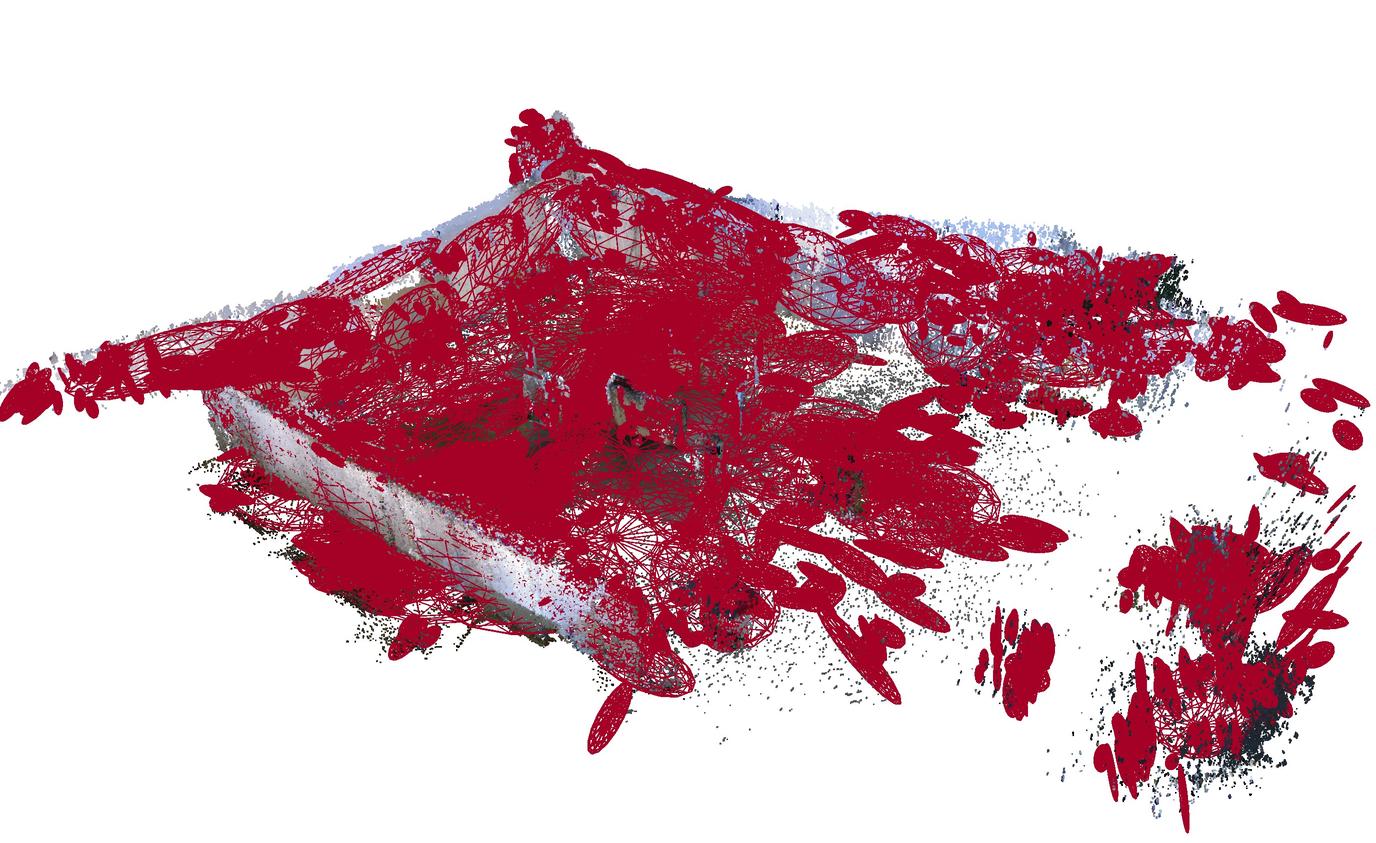}
		\label{fig:warehouse_gau}
	}
	\hfill
	\subfloat[Occupancy distribution at a cross section]{%
		\includegraphics[width=0.29\textwidth]{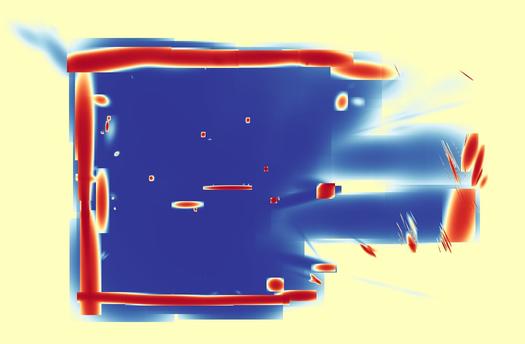}
		\label{fig:warehouse_occ}
	}
	\caption{Visualization of (a) the point cloud overlaid with its (b) GMMap for \emph{Warehouse} (structured indoor) environment. For ease of visualization, only occupied Gaussians are shown for the GMMap. 
	In (c), the distribution of occupancy at a horizontal cross section of the GMMap is visualized in the free regions (blue), unexplored regions (yellow), and occupied regions (red).
	The locations of the unexplored regions are well-preserved. \plcomment{Depending on the application requirements, the unexplored prior weight ($\pi_0$) of the GMMap can be used to increase (\emph{i.e.}, smaller $\pi_0$) or decrease (\emph{i.e.}, larger $\pi_0$) the extent of free regions interpolated into the unexplored region during Gaussian Mixture Regression.}}
	\label{fig:map_viz_warehouse} 
\end{figure*}

\begin{figure*}
	\centering
	\subfloat[\emph{Soulcity} point cloud]{%
		\includegraphics[width=0.32\textwidth]{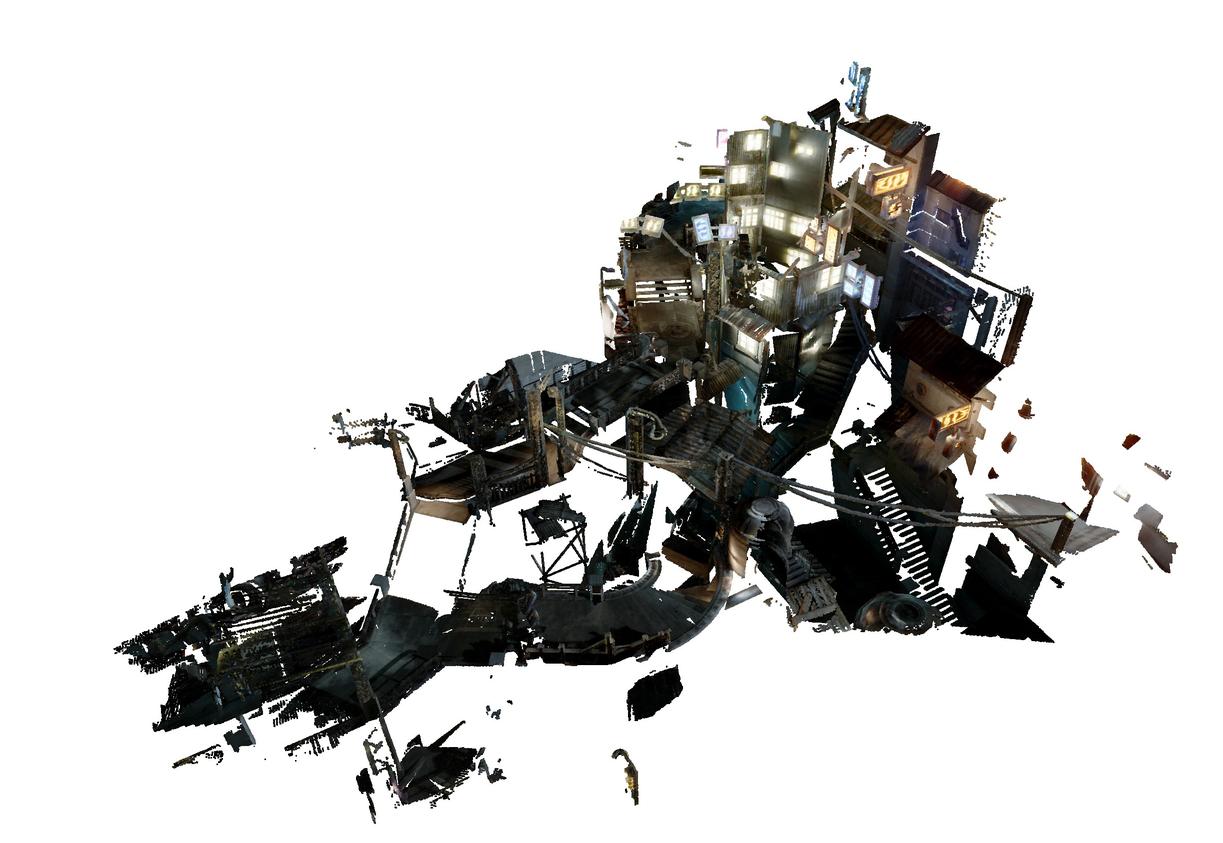}
		\label{fig:soulcity_pcd}
	}
	\hfill
	\subfloat[GMMap]{%
		\includegraphics[width=0.32\textwidth]{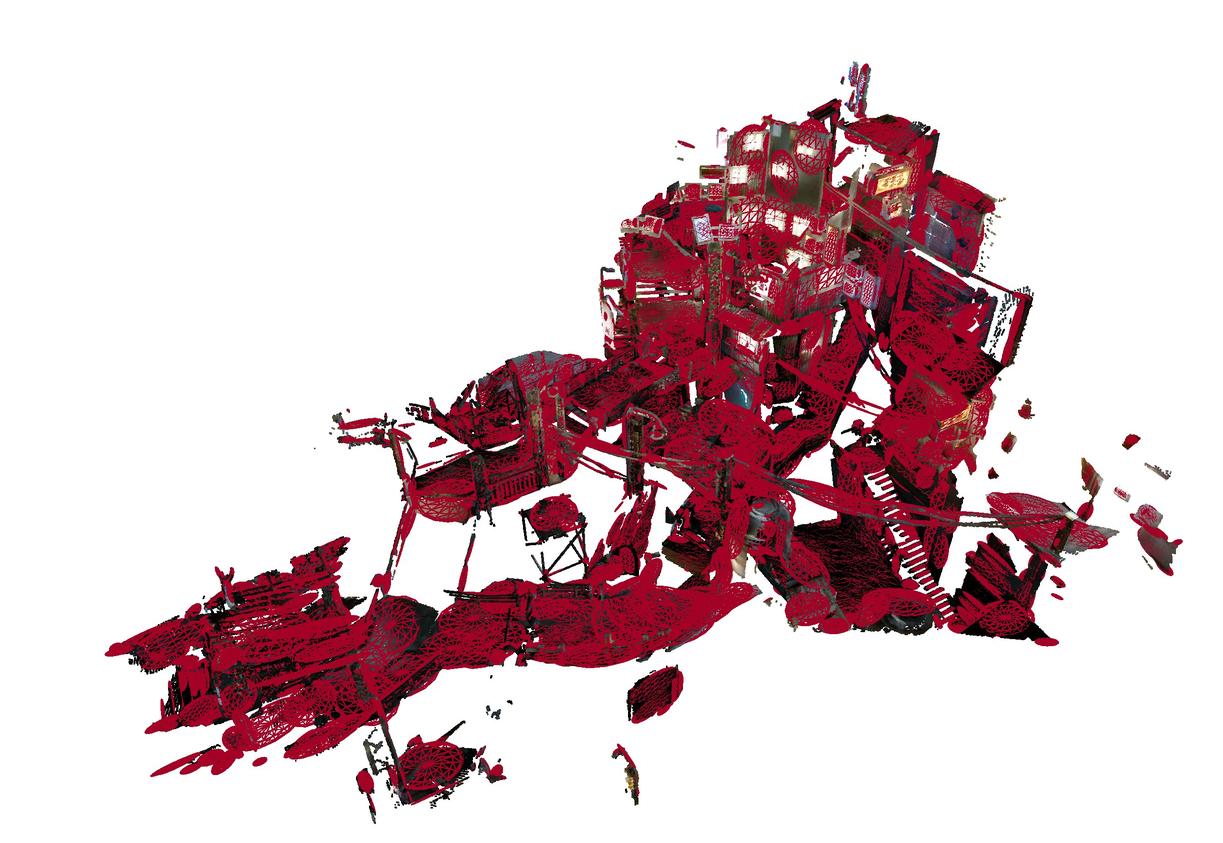}
		\label{fig:soulcity_gau}
	}
	\hfill
	\subfloat[Occupancy distribution at a cross section]{%
		\includegraphics[width=0.29\textwidth]{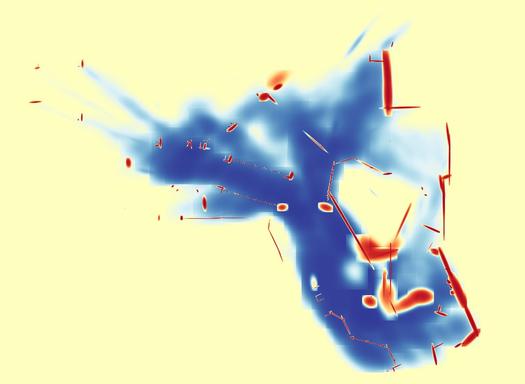}
		\label{fig:soulcity_occ}
	}
	
	\caption{Visualization of (a) the point cloud overlaid with its (b) GMMap for \emph{Soulcity} (structured outdoor) environment. For ease of visualization, only occupied Gaussians are shown for the GMMap. 
		In (c), the distribution of occupancy at a horizontal cross section of the GMMap is visualized in the free regions (blue), unexplored regions (yellow), and occupied regions (red).
		The locations of the unexplored regions are well-preserved. \plcomment{Depending on the application requirements, the unexplored prior weight ($\pi_0$) of the GMMap can be used to increase (\emph{i.e.}, smaller $\pi_0$) or decrease (\emph{i.e.}, larger $\pi_0$) the extent of free regions interpolated into the unexplored region during Gaussian Mixture Regression.}}
	\label{fig:map_viz_soulcity} 
\end{figure*}

The visualization of GMMap and the preservation of unexplored regions for \emph{Warehouse} and \emph{Soulcity} are illustrated in \fig~\ref{fig:map_viz_warehouse} and \ref{fig:map_viz_soulcity}, respectively.
\plcomment{Unlike discrete representations that assume occupancy is spatially independent, Gaussians allow the GMMap to spatially interpolate occupancy to achieve higher compactness. 
Thus, another fundamental trade-off of the GMMap is the total extent of free regions classified for safe navigation \emph{v.s.} the extent of free regions interpolated by free Gaussians into the unexplored region at the frontiers (\emph{i.e.}, boundaries between free and unexplored regions).
This trade-off can be controlled using the unexplored prior weight ($\pi_0$) that dictates the contribution of unexplored regions during GMR.
%
Specifically, if the application is more sensitive to the exact locations of the frontier, the prior weight ($\pi_0$) can be increased to reduce the total extent of free regions classified by GMMap for safe traversal.
Even if the prior weight ($\pi_0$) is unchanged, the frontiers (and regions near obstacle surfaces) can be identified at locations of high occupancy variance computed using \eqn~\eqref{eqn:occ_variance}.}

\subsection{Construction \& Query Throughput}~\label{subsec:throughput}
In this section, we compare the computational efficiency of our GMMap against other frameworks using the NVIDIA Jetson TX2 platform.
The computational efficiency is evaluated in terms of the throughput for constructing the map (\emph{i.e.}, depth images per second, in Section~\ref{subsubsec:tp_map}) and also querying the map (\emph{i.e.}, locations per second, in Section~\ref{subsubsec:tp_query}).
\tab~\ref{tab:comp_summary} summarizes these metrics for all frameworks.

\subsubsection{\textbf{Map Construction}}~\label{subsubsec:tp_map}
The NVIDIA Jetson TX2 platform contains a low-power ARM Cortex A57 CPU with four cores and a Pascal GPU with two Streaming Multiprocessors (SMs).
Due to computationally efficient GMM generation and fusion, our GMMap can be constructed at a throughput of 11 to 18 images per second using only one CPU core, which is $4\times$ to $36\times$ higher than other frameworks.
Since Scanline Segmentation (\algline~\ref{spgf:scan_seg} in \alg~\ref{alg:gmm_creation}) dominates the amount of computation during map construction and can be concurrently executed across multiple rows of the depth image, our construction throughput can be significantly increased via parallelization. 
By using all four CPU cores, our multi-core implementation reaches a throughput of 31 to 60 images per second. 
Multi-core implementations of existing frameworks are either not publicly available or highly experimental. Even if these frameworks can be effectively parallelized with four cores, their throughputs are expected to be $4 \times$ higher, which are still much lower than our multi-core implementation.
%
By concurrently executing Scanline Segmentation across four images, our GPU implementation of GMMap offers the highest construction throughput of 44 to 81 images per second, which is up to $2\times$ higher\footnote{Even though we are processing four images at the same time, the throughput is not four times higher because other sequential procedures of the GMMap construction (\emph{i.e.}, segment and GMM fusion) start to dominate.} than our CPU multi-core implementation.

\subsubsection{\textbf{Occupancy Query}}~\label{subsubsec:tp_query}
\tab~\ref{tab:comp_summary} also compares the query throughput of our GMMap against existing frameworks. To emulate an energy-constrained setting during path planning, each map is queried at locations throughout all observed regions (\emph{i.e.}, no unexplored regions) in the environment using only a single CPU core.
Recall that each map consists of geometric primitives (\emph{e.g.}, Gaussians or voxels) stored using a spatial data structure (\emph{e.g.}, grid, R-tree, or octree).
%
%
For existing frameworks, either traversing the spatial data structure (\emph{i.e.}, accessing a voxel from a grid in NDT-OM) or inferring occupancy from primitives (\emph{i.e.}, reading occupancy probability in BGKOctoMap-L and OctoMap) require little compute, which leads to high query throughputs ranging from $9.3 \times 10^5$ to $4.2 \times 10^6$ locations per second.
However, in our GMMap, both R-tree traversal and Gaussian Mixture Regression (GMR) require more computation.
Thus, the query throughput is lower than other frameworks but still sufficiently high (ranging from $4.6 \times 10^5$ to $7.9 \times 10^5$ locations per second).
If needed, the query throughput can be increased by accessing the map with multiple cores and/or partitioning the query locations into more localized sets for batch processing.

\begin{table*}[!t]
	\centering
	\caption{Comparison of the GMMap against prior works using the NVIDIA Jetson TX2
	}
	\resizebox{0.98\textwidth}{!}{
		\begin{threeparttable}
		\begin{tabular}{ccccccccccc} 
			\toprule
			\multirow{3}{*}{\textbf{Environment}} & \multirow{3}{*}{\textbf{Framework}} & \multirow{2}{*}{\makecell{\textbf{Compute} \\ \textbf{Resource}}} & \multicolumn{2}{c}{\textbf{Throughput}} & \multicolumn{3}{c}{\textbf{Memory Footprint}} & \multicolumn{3}{c}{\textbf{Energy Consumption}} \\
			
			& & & \makecell{Construction} & \makecell{Query\tnote{*}} & \makecell{Map Size} & \makecell{Overhead} & \makecell{DRAM Access}  & \makecell{CPU \& GPU} & \makecell{DRAM} & \makecell{Total} \\
			
			& & (C = CPU core) &\makecell{(images/s)} & \makecell{($10^6$ locations/s)} & \makecell{(KB)} & \makecell{(KB)} & \makecell{(bytes/pixel)} & \makecell{(mJ/image)} & \makecell{(mJ/image)} & \makecell{(mJ/image)} \\
			\midrule 
			
			\multirow{6}{*}{Room} & \multirow{3}{*}{GMMap} & GPU \& 4 C & $\mathbf{81}$ & $0.79$ & $\mathbf{167}$ & $24,563$\tnote{**} & $477$ & $41$ & $17$ & $58$ \\
			
			&  & 4 C & $60$ & $0.79$ & $176$ & $41$ & $27$ & $\mathbf{36}$ & $\mathbf{16}$ & $\mathbf{52}$ \\
			
			&  & 1 C & $18$ & $0.79$ & $176$ & $\mathbf{31}$ & $\mathbf{14}$ & $59$ & $51$ & $110$ \\
			
			& NDT-OM & 1 C & $5.0$ & $3.5$ & $426$ & $3,146$ & $160$  & $202$ & $157$ & $359$ \\			 
			
			& BGKOctoMap-L & 1 C & $2.8$ & $0.93$ & $4,935$ & $7,101$ & $242$ & $352$ & $272$ & $624$ \\
			
			& OctoMap & 1 C & $3.6$ & $\mathbf{4.0}$ & $2,190$ & $629$ & $164$ & $298$ & $209$ & $507$ \\
			
			\midrule
			
			\multirow{6}{*}{Warehouse} & \multirow{3}{*}{GMMap} & GPU \& 4 C & $\mathbf{73}$ & $0.52$ & $\mathbf{268}$ & $24,596$\tnote{**} & $492$ & $43$ & $16$ & $59$ \\
			
			&  & 4 C & $58$ & $0.51$ & $269$ & $56$ & $30$ & $\mathbf{37}$ & $\mathbf{14}$ & $\mathbf{51}$ \\
			
			&  & 1 C& $18$ & $0.51$ & $269$ & $\mathbf{41}$ & $\mathbf{20}$ & $59$ & $41$ & $100$ \\
			
			& NDT-OM & 1 C& $3.7$ & $3.7$ & $614$ & $3,436$ & $199$ & $273$ & $209$ & $482$ \\		
			
			& BGKOctoMap-L & 1 C & $0.5$ & $1.4$ & $13,811$ & $21,265$ & $940$  & $1,888$ & $1,463$ & $3,351$ \\
			
			& OctoMap & 1 C & $4.3$ & $\mathbf{4.2}$ & $1,590$ & $606$ & $143$ & $256$ & $176$ & $433$ \\
			
			\midrule
			
			\multirow{6}{*}{Soulcity} & \multirow{3}{*}{GMMap} & GPU \& 4 C & $\mathbf{60}$ & $0.46$ & $850$ & $24,740$\tnote{**} & $625$ & $\mathbf{56}$ & $\mathbf{23}$ & $\mathbf{79}$ \\
			
			&  & 4 C & $31$ & $0.47$ & $\mathbf{838}$ & $128$ & $76$  & $73$ & $25$ & $98$ \\
			
			&  & 1 C & $11$ & $0.47$ & $\mathbf{838}$ & $\mathbf{106}$ & $\mathbf{44}$ & $92$ & $66$ & $158$ \\
			
			& NDT-OM & 1 C & $3.1$ & $3.9$ & $1,925$ & $4,391$ & $372$  & $324$ & $248$ & $572$ \\			
			
			& BGKOctoMap-L & 1 C & $0.8$ & $1.0$ & $23,265$ & $5,502$ & $596$  & $1,204$ & $926$ & $2,130$ \\
			
			& OctoMap & 1 C & $2.1$ & $\mathbf{4.1}$ & $10,452$ & $1,068$ & $644$ & $485$ & $373$ & $858$ \\
			
			\midrule
			
			\multirow{6}{*}{Gascola} & \multirow{3}{*}{GMMap} & GPU \& 4 C & $\mathbf{44}$ & $0.62$ & $362$ & $24,644$\tnote{**} & $1,048$  & $\mathbf{69}$ & $36$ & $105$ \\
			
			&  & 4 C & $32$ & $0.62$ & $\mathbf{361}$ & $79$ & $78$  & $73$ & $\mathbf{29}$ & $\mathbf{102}$ \\
			
			&  & 1 C & $11$ & $0.62$ & $\mathbf{361}$ & $\mathbf{63}$ & $\mathbf{54}$ & $97$ & $81$ & $178$ \\
			
			& NDT-OM & 1 C & $2.6$ & $\mathbf{3.9}$ & $1,339$ & $4,392$ & $358$ & $383$ & $291$ & $674$ \\			
			
			& BGKOctoMap-L & 1 C & $0.4$ & $1.1$ & $16,736$ & $9,993$ & $899$  & $2,407$ & $1,840$ & $4,248$ \\
			
			& OctoMap & 1 C & $1.6$ & $\mathbf{3.9}$ & $9,376$ & $760$ & $1,136$ & $634$ & $494$ & $1,129$ \\			
			
			\bottomrule
		\end{tabular}
		\begin{tablenotes}
			\item[*] Unlike other metrics, query throughput is computed using a single CPU core.
			\item[**] High memory overhead due to the necessary allocation of large GPU-accessible buffers (used to store input images and output results of Scanline Segmentation) for concurrent processing of four images. Allocations of these buffers are not required for CPU-only implementations. All frameworks achieve comparable accuracy. Bold entity achieves the best performance in its corresponding metric.
		\end{tablenotes}
		\end{threeparttable}
		\label{tab:comp_summary}
	}
\end{table*}

\subsection{Memory Footprint}~\label{subsec:memory}
In this section, we compare the memory efficiency of our GMMap against other frameworks when executing on the NVIDIA Jetson TX2 platform.
In addition to the map size (Section~\ref{subsubsec:memory_map_size}), we are interested in the memory overhead (for storing input and temporary variables, in Section~\ref{subsubsec:memory_overhead}) and the amount of DRAM access per pixel (which dictates DRAM energy consumption, in Section~\ref{subsubsec:memory_dram}) during the map construction.
\tab~\ref{tab:comp_summary} summarizes our results.

\subsubsection{\textbf{Map Size}}~\label{subsubsec:memory_map_size}
We compare the size of the map that includes the geometric primitives (\emph{e.g.}, Gaussians and/or voxels) and the spatial data structure (\emph{i.e.}, R-tree, grid and/or octree) among all frameworks.
Due to the compactness and strong representational power of the Gaussians, NDT-OM achieves comparable accuracy while reducing the map size by 61\% to 96\% compared with BGKOctoMap-L and OctoMap. 
However, the extent of each Gaussian in NDT-OM is restricted by the constant voxel size across the entire environment. Thus, all Gaussians appear similarly sized as shown in \fig~\ref{fig:ndt_room} and \ref{fig:ndt_gascola}.
By using SPGF* to construct Gaussians that appropriately adapt to the geometries of occupied and free regions in the environment (see \fig~\ref{fig:gmmap_room} and \ref{fig:gmmap_gascola}), our GMMap achieves comparable accuracy while reducing the map size by 56\% to 73\% compared with NDT-OM.
Across all frameworks, GMMap requires the least amount of memory (167KB to 850KB) across all four environments.

\subsubsection{\textbf{Memory Overhead}}~\label{subsubsec:memory_overhead}
We are interested in the memory overhead\footnote{\plcomment{Measured using our own memory profiler that automatically tracks memory allocation in the constructors and deconstructors of relevant C++ objects.}} (defined as the peak memory usage minus the map size) for storing input and temporary variables during map construction.
For a memory-efficient framework, its memory overhead should be insignificant compared with the map size.
Unfortunately, existing frameworks are not memory efficient.
For NDT-OM, the memory overhead mostly comprises the point cloud associated with each depth image (up to 3.6MB) for supporting a variety of edge cases during recency-weighted covariance update~\cite{saarinen20133d}.
For BGKOctoMap-L, the memory overhead mostly is comprised of subsampled measurements in free and occupied regions (up to 21MB) for performing multi-pass BGK inference.
For OctoMap, the memory overhead is mostly comprised of pointers to a large number of voxels intersected by sensor rays from each depth image (up to 1MB).

In contrast, our GMMap requires very little memory overhead.
Since SPGF* processes the depth image one pixel at a time in a single pass, only one (for single-core implementation) or four pixels (for multi-core implementation) are stored in memory at any time.
Thus, the memory overhead associated with map construction is mostly comprised of compact line segments $\dist{S}$ generated from Scanline Segmentation in SPGF* and the local GMMap $\dist{G}_t$ generated at the output of \alg~\ref{alg:gmm_creation}. 
From \tab~\ref{tab:comp_summary}, the memory overhead of our single-core implementation is only 31KB to 106KB, which is at least 90\% lower than other frameworks.
Since four scanlines are segmented concurrently in our multi-core implementation, the memory overhead increases and ranges from 41KB to 128KB, which is at least 88\% lower than other frameworks.
However, our GPU implementation requires much larger memory overhead (around 24MB) due to the allocations of large GPU-accessible buffers for transferring four depth images and their Scanline Segmentation outputs to and from the GPU.

\subsubsection{\textbf{DRAM Access}}~\label{subsubsec:memory_dram}
We compare the average amount of DRAM access required for integrating each measurement (\emph{i.e.}, pixel in the depth image) into the map among all frameworks.
The amount of DRAM access correlates with the energy consumption of the DRAM and is computed by multiplying the number of last-level cache misses\footnote{\plcomment{Obtained by reading hardware counters of the CPU using the $\mathrm{perf\_event\_open()}$ system call on Linux. For GPU, the amount of DRAM accesses can be obtained directly using NVIDIA NSight Systems.}} with the size of the cache line.
Recall that existing frameworks update the map by incrementally casting each measurement ray (more than 300,000 rays in a 640$\times$480 depth image) into the current map.
Since these rays diverge away from the sensor origin, memory accesses along these rays often lack spatial and temporal locality for effective cache usage (especially if the map is too large to fit within on-chip caches).
Thus, the single-core implementations of existing frameworks require a significant number of DRAM accesses ranging from 160 bytes to more than 1KB per pixel.

In contrast, our GMMap avoids ray casting by directly fusing Gaussians from a compact local map $\dist{G}_t$ with Gaussians from the previously observed region $\dist{C}_t$ in the global map $\dist{M}_{t-1}$ (see \fig~\ref{fig:gmm_fusion}).
Since both the local map $\dist{G}_t$ and the previously observed region $\dist{C}_t$ can be compactly cached, our single-core implementation reduces DRAM access by at least 85\% (compared to existing frameworks) by accessing only 14 bytes to 54 bytes per pixel.
Since multiple cores share the last-level cache, the number of cache misses increases for our multi-core implementation which requires slightly higher DRAM accesses ranging from 27 bytes to 78 bytes per pixel (at least 78\% lower than existing frameworks).
Our GPU implementation requires much larger DRAM accesses due to the higher amount of cache misses from the concurrent segmentation of all scanlines in four images.
However, most DRAM accesses from our GPU implementation are coalesced (\emph{i.e.}, multiple accesses can be serviced with a single transaction). Thus, the energy consumption of the DRAM slightly increases compared with our multi-core CPU implementation.

\subsection{Energy Consumption}~\label{subsec:energy_consumption}
\tab~\ref{tab:comp_summary} summarizes the average energy consumption\footnote{\plcomment{Measured directly from the power monitors embedded on the Jetson TX2.}} per depth image during map construction.
For all frameworks, the energy consumption of the DRAM is significant compared with that of CPU and GPU, which underscores the importance of reducing memory overhead and access.
Due to computationally efficient single-pass GMM creation and fusion, our single-core implementation reduces the energy consumption of the CPU by at least 71\% compared with other frameworks.
By avoiding ray casting (and its associated DRAM accesses), our single-core implementation reduces the energy consumption of the DRAM by at least 68\% compared with other frameworks.
Our multi-core and GPU implementations are even more energy efficient. \plcomment{Recall that energy equals the product of power and latency. For both implementations, the decrease in average latency per image significantly outweighs the increase in power consumption. Thus, the energy consumption per image decreases compared with single-core implementation.}
In all, our CPU single-core, multi-core, and GPU implementations reduce the average energy consumption by at least 69\%, 83\%, and 84\% compared with other frameworks, respectively.

\section{Conclusion}
In this article, we proposed the GMMap that uses a compact Gaussian mixture model to accurately model the continuous distribution of occupancy in 3D environments.
Occupancy probability is inferred with Gaussian Mixture Regression which is extended to retain unexplored regions.
\plcomment{Across multiple indoor and outdoor environments, we analyzed GMMap's ability to accurately model free/occupied regions as well as boundaries from free-to-obstacle, and free-to-unexplored regions. Furthermore, we explained how to tune the parameters of the GMMap for controlling two fundamental trade-offs (\emph{i.e.}, modeling of distant small objects \emph{v.s.} compactness of the map, and the total extent of free regions \emph{v.s.} extension of free regions into unexplored regions) to ensure safe navigation of robots.}

Due to single-pass processing of depth images and directly operating on Gaussians, GMMap can be constructed in real-time on energy-constrained platforms while significantly reducing memory overhead and access.
When benchmarked on the low-power NVIDIA Jetson TX2 platform across a diverse set of environments, GMMap can be constructed at a throughput of up to 60 images per second using the CPU and up to 81 images per second using the GPU, which is $4 \times$ to $146 \times$ higher than prior works.
While achieving comparable accuracy as prior works, our CPU implementation of GMMap reduces map size by at least 56\%, memory overhead by at least 88\%, DRAM access by at least 78\%, and energy consumption by at least 69\%.
Thus, to the best of our knowledge, GMMap not only enables real-time large-scale 3D mapping for energy-constrained robots for the first time but also illustrates the significance of memory-efficient algorithms for enabling low-power autonomy on these robots.




\bibliographystyle{IEEEtran}
\bibliography{references}
\section{Biography}
\vspace{-33pt}
\begin{IEEEbiography}[{\includegraphics[width=1in,height=1.25in,clip,keepaspectratio]{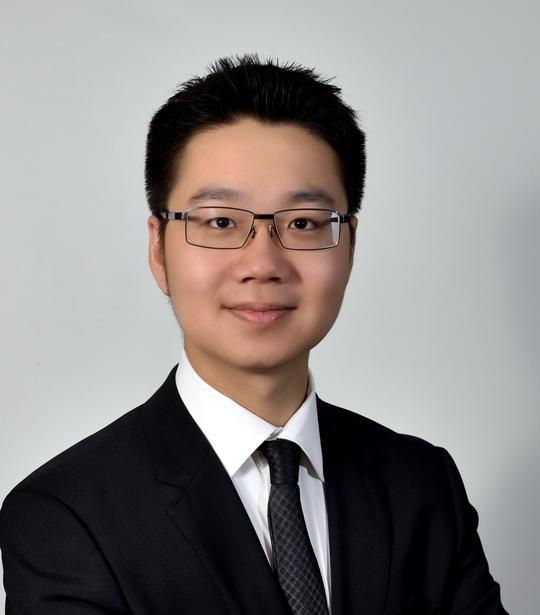}}]{Peter Zhi Xuan Li} (Student Member, IEEE) received the B.A.Sc. in Engineering Science from the University of Toronto, Canada, in 2018. Between 2016 and 2017, he worked in the High-Speed Converters Group at Analog Devices, Toronto, as an integrated circuit engineer. His research focuses on the co-design of memory-efficient algorithms and specialized hardware for localization, mapping, and path-planning on energy-constrained devices such as AR/VR headsets, smartphones, and micro-robots.
\end{IEEEbiography}

\vskip -2\baselineskip plus -1fil
\begin{IEEEbiography}[{\includegraphics[width=1in,height=1.25in,clip,keepaspectratio]{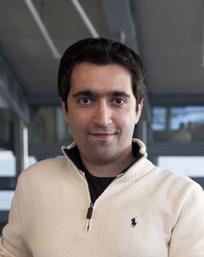}}]{Sertac Karaman} (Member, IEEE) received the B.S. degrees in mechanical engineering and computer engineering from the Istanbul Technical University,
Istanbul, Turkey, in 2007, the S.M. degree in mechanical engineering and the Ph.D. degree in electrical engineering and computer science from the Massachusetts Institute of Technology (MIT), Cambridge, MA, USA, in 2009 and 2012, respectively.
He is currently a Professor of Aeronautics and Astronautics at MIT. His research interests include the broad areas of robotics and control theory. In particular, he is focusing on the applications of probability theory, stochastic processes, stochastic geometry, formal methods, and optimization for the design and analysis of high-performance cyber-physical systems. The application areas of his research include driverless cars, unmanned aerial vehicles, distributed aerial surveillance systems, air traffic control, certification and verification of control systems software, and many others.

Dr. Karaman was the recipient of the IEEE Robotics and Automation Society Early Career Award, in 2017, the Office of Naval Research Young Investigator Award, in 2017, the Army Research Office Young Investigator Award, in 2015, the National Science Foundation Faculty Career Development (CAREER) Award, in 2014, the AIAA Wright Brothers Graduate Award, in 2012, and the NVIDIA Fellowship, in 2011.
\end{IEEEbiography}

\vskip -2\baselineskip plus -1fil
\begin{IEEEbiography}[{\includegraphics[width=1in,height=1.25in,clip,keepaspectratio]{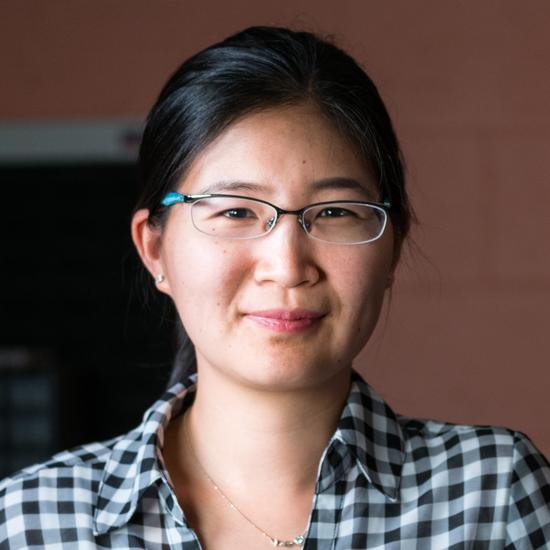}}]{Vivienne Sze} (Senior Member, IEEE) received the B.A.Sc. (Hons) degree in electrical engineering from the University of Toronto, Toronto, ON, Canada, in 2004, and the S.M. and Ph.D. degree in electrical engineering from the Massachusetts Institute of Technology (MIT), Cambridge, MA, in 2006 and 2010 respectively. In 2011, she received the Jin-Au Kong Outstanding Doctoral Thesis Prize in Electrical Engineering at MIT. 

She is an Associate Professor at MIT in the Electrical Engineering and Computer Science Department. Her research interests include computing systems that enable energy-efficient machine learning, computer vision, and video compression/processing for various applications, including autonomous navigation, digital health, and the Internet of Things.  Prior to joining MIT, she was a Member of the Technical Staff in the Systems and Applications R\&D Center at Texas Instruments (TI), Dallas, TX, where she designed low-power algorithms and architectures for video coding. She also represented TI in the Joint Collaborative Team on Video Coding (JCT-VC).

Dr. Sze was a recipient of the Air Force Young Investigator Research Program Award, the DARPA Young Faculty Award, the Edgerton Faculty Award, several faculty awards from Google, Facebook, and Qualcomm, the 2021 University of Toronto Engineering Mid-Career Achievement Award, and the 2020 ACM-W Rising Star Award, and a co-recipient of the 2018 Symposium on VLSI Circuits Best Student Paper Award, the 2017 CICC Outstanding Invited Paper Award, and the 2016 IEEE Micro Top Picks Award. She was a member of the JCT-VC team that received the Primetime Engineering Emmy Award for the development of the HEVC video compression standard.

\end{IEEEbiography}

\vfill

\end{document}